\documentclass[10pt,twocolumn,letterpaper]{article}

\usepackage[pagenumbers]{cvpr} 
\usepackage[accsupp]{axessibility}  

\usepackage{lipsum}
\usepackage{array}
\usepackage{times}
\usepackage{epsfig}
\usepackage{graphicx}
\usepackage{float}
\usepackage{wrapfig}
\usepackage{amsmath,amssymb,amsthm}
\usepackage{algorithm,algorithmicx,algpseudocode}
\usepackage{bm,xspace}
\usepackage{comment}
\usepackage{multirow}
\usepackage{balance}
\usepackage{url}
\usepackage{booktabs}
\usepackage{etoolbox,siunitx}
\usepackage{calc}
\usepackage{pifont,hologo}
\usepackage{color}
\usepackage{adjustbox}
\usepackage{amsmath}
\usepackage{enumitem}
\usepackage{bbding}  %
\usepackage[normalem]{ulem}  %
\usepackage{contour}
\usepackage{soul}%
\usepackage{tocloft} %
\usepackage{etoc} %
\definecolor{cvprblue}{rgb}{0.21,0.49,0.74}
\usepackage[pagebackref,breaklinks,colorlinks,allcolors=cvprblue]{hyperref}
\usepackage{natbib}
\PassOptionsToPackage{square,numbers,comma,sort}{natbib}

\definecolor{blue}{HTML}{004bb3}
\definecolor{red}{HTML}{cc1100}
\definecolor{orange}{HTML}{cc7700}
\definecolor{gray}{HTML}{efefef}
\definecolor{darkgreen}{HTML}{228B22}
\definecolor{darkgray}{HTML}{757575}

\definecolor{cite}{HTML}{3270b5}
\definecolor{link}{HTML}{b53532}
\definecolor{link}{HTML}{cc1100}
\definecolor{normal}{HTML}{001219}
\definecolor{tta}{HTML}{0077BE}
\definecolor{peft}{HTML}{FFA07A}
\definecolor{seablue}{RGB}{70, 130, 180}
\definecolor{lightorange}{RGB}{255, 180, 140}

\newcommand{\peft}{\textcolor{peft}{$\mathbf{\circ}$\,}}
\newcommand{\normal}{\textcolor{normal}{$\mathbf{\circ}$\,}}   
\newcommand{\tta}{\textcolor{tta}{$\bullet$\,}}

\contourlength{0.8pt}

\renewcommand{\eqref}[1]{Eq.~\ref{#1}}

\newcolumntype{x}[1]{>{\centering\arraybackslash}p{#1}}
\newcolumntype{y}[1]{>{\raggedright\arraybackslash}p{#1}}
\newcolumntype{z}[1]{>{\raggedleft\arraybackslash}p{#1}}
\newcommand{\tablestyle}[2]{\setlength{\tabcolsep}{#1}\renewcommand{\arraystretch}{#2}\centering\footnotesize}

\DeclareMathSymbol{@}{\mathord}{letters}{"3B}

\makeatletter
\DeclareRobustCommand\onedot{\futurelet\@let@token\@onedot}
\def\@onedot{\ifx\@let@token.\else.\null\fi\xspace}

\newcommand*{\Rom}[1]{\expandafter\@slowromancap\romannumeral #1@}
\newcommand*{\rom}[1]{\expandafter\romannumeral #1}

\def\1{\bm{1}}

\DeclareMathAlphabet{\mathsfit}{\encodingdefault}{\sfdefault}{m}{sl}
\SetMathAlphabet{\mathsfit}{bold}{\encodingdefault}{\sfdefault}{bx}{n}

\let\originalleft\left
\let\originalright\right
\renewcommand{\left}{\mathopen{}\mathclose\bgroup\originalleft}
\renewcommand{\right}{\aftergroup\egroup\originalright}

\def\paperID{8563} 
\def\confName{CVPR}
\def\confYear{2025}

\begin{document}

\title{SoMA: Singular Value Decomposed Minor Components Adaptation \\ for Domain Generalizable Representation Learning}

\author{Seokju Yun\qquad Seunghye Chae\qquad Dongheon Lee\qquad  Youngmin Ro\textsuperscript{*} \\Machine Intelligence Laboratory, University of Seoul, Korea \\ \tt\small \{wsz871, tmdhey, dslisleedh, youngmin.ro\}@uos.ac.kr \\ {\tt\small \url{https://github.com/ysj9909/SoMA}}}

\twocolumn[{%
\renewcommand\twocolumn[1][]{#1}%
\vspace{-13mm}
\maketitle
\vspace{-11mm}
\begin{center}
    \captionsetup{type=figure}
    \includegraphics[width=\linewidth]{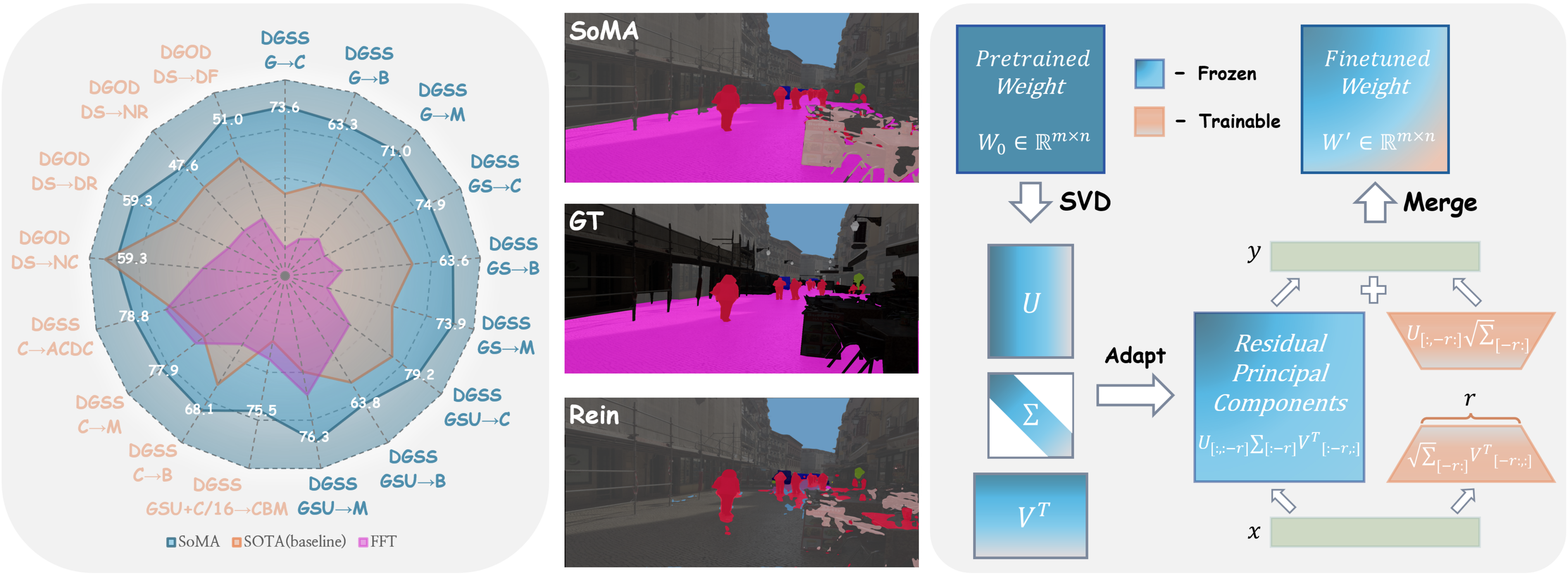}
    \vspace{-6mm}
    \captionof{figure}{\textbf{Overview of SoMA framework.} \emph{Left.} 
    SoMA achieves state-of-the-art results across diverse tasks, ranging from domain-generalized semantic segmentation (DGSS) to object detection (DGOD), and performs well in both {\color{seablue} synthetic-to-real} and {\color{lightorange} real-to-real} scenarios.
    \emph{Middle.} Our method, trained solely on synthetic datasets, demonstrates strong generalization capabilities in complex real-world scenes.
    \emph{Right.} We present SoMA, a method that applies SVD to the pre-trained weights, decomposing them into $r$ minor singular components $U_{[:,-r:]}\Sigma_{[-r:]} (V^T)_{[-r:,:]}$ and residuals.
    SoMA selectively tunes the smallest $r$ components, effectively preserving world knowledge of foundation models. Since SoMA shares the same forward architecture as LoRA~\cite{hu2022lora}, it adds no extra latency during inference phase.
    }\label{fig:teaser}
\end{center}%
}]
\begin{abstract}
\renewcommand{\thefootnote}{\fnsymbol{footnote}}
\footnotetext[1]{Corresponding author.}
    Domain generalization (DG) aims to adapt a model using one or multiple source domains to ensure robust performance in unseen target domains.
Recently, Parameter-Efficient Fine-Tuning (PEFT) of foundation models has shown promising results in the context of DG problem.
Nevertheless, existing PEFT methods still struggle to strike a balance between preserving generalizable components of the pre-trained model and learning task-specific features.
To gain insights into the distribution of generalizable components, we begin by analyzing the pre-trained weights  through the lens of singular value decomposition.
Building on these insights, we introduce \textbf{S}ingular Value Dec\textbf{o}mposed \textbf{M}inor Components \textbf{A}daptation (\textbf{SoMA}), an approach that selectively tunes minor singular components while keeping the residual parts frozen.
SoMA effectively retains the generalization ability of the pre-trained model while efficiently acquiring task-specific skills.
Moreover, we freeze domain-generalizable blocks and employ an annealing weight decay strategy, thereby achieving an optimal balance in the delicate trade-off between generalizability and discriminability. 
SoMA attains state-of-the-art results on multiple benchmarks that span both domain generalized semantic segmentation to domain generalized object detection.
In addition, our methods introduce no additional inference overhead or regularization loss, maintain compatibility with any backbone or head, and are designed to be versatile, allowing easy integration into a wide range of tasks.
   
\vspace{-0.5mm}
\end{abstract}

\vspace{-6mm}
\section{Introduction} \label{sec:intro}
\vspace{-1mm}
Deep neural networks (DNNs) have recently made significant strides in dense prediction tasks, including semantic segmentation~\cite{cheng2022mask2former} and object detection~\cite{detr, ren2015fasterrcnn}, both of which are crucial for safety-critical applications, such as autonomous driving.
However, DNNs often fail to maintain reliable performance under domain shifts, which may result from diverse lighting conditions, weather changes, or differences in location.
Likewise, while synthetic data~\cite{gtav, synthia, urbansyn} is employed to avoid the labor-intensive and costly process of constructing real-world datasets, the resulting performance degradation of DNNs in real-world deployment remains a critical issue to be resolved.
To tackle these challenges, Domain Generalization (DG) is introduced to design models capable of consistent prediction on arbitrary unseen target domains.

Existing DG methods for dense prediction tasks employ two main strategies to enhance model robustness.
The first involves \textbf{diversify}ing the training process in the input~\cite{chattopadhyay2023pasta, peng2021globallocalTR} or feature space~\cite{udupa2024mrfp, NP} using augmentation/perturbation techniques, exposing the model to a broader range of styles to mitigate overfitting to specific domains.
The second strategy focuses on imposing \textbf{align}ment constraints, such as normalization~\cite{sansaw} or regularization losses~\cite{choi2021robustnet, ahn2024blindnet}, to facilitate the learning of domain-invariant features.
However, these methods mostly utilize outdated backbones (\textit{e.g.}, ResNet~\cite{resnet} or MobileNetV2~\cite{sandler2018mobilenetv2}) pre-trained on mid-scale datasets, such as ImageNet~\cite{deng2009imagenet}.

Meanwhile, vision foundation models (VFMs)~\cite{awais2023foundational}, pre-trained on vast, curated datasets with enormous parameter counts, have recently shown unprecedented generalization capabilities~\cite{rein}.
Given this landscape, we propose a paradigm shift—from applying the \textbf{diversify \& align} approach to classic backbones, to devising methods that \textbf{preserve the world knowledge of VFMs while effectively learning task-specific features}.
However, naive adaptation of VFMs via \textit{full fine-tuning} (FFT), which involves retraining all model parameters, results in prohibitive costs and a risk of catastrophic forgetting of the pre-trained knowledge.

In light of these challenges, a commonly adopted approach is parameter-efficient fine-tuning (PEFT), where only the injected lightweight adapters~\cite{chen2022adaptformer, unifiedviewadapter, hu2022lora} or tokens~\cite{jia2022vpt, lester-etal-2021-power} are fine-tuned, while the pre-trained weights remain frozen.
Low-Rank Adaptation (LoRA)~\cite{hu2022lora} is one of the most widely used PEFT methods, which substitutes the updates with the product of two low-rank matrices.
Interestingly, we found that LoRA exhibits superior generalization ability in DG dense prediction tasks compared to FFT and other PEFT methods (see Tab.~\ref{tab: dgss_s2r}), aligning with recent findings in LoRA-based literature~\cite{biderman2024loraforgetless, pego}.
Nevertheless, LoRA adapters are built without considering the distribution of generalizable components within the pre-trained weights they modify.
As a result, there is still room for enhancing generalizability by preserving the diverse domain knowledge of VFMs.

To uncover the structure in which world knowledge is encoded within the pre-trained weights, we first perform singular value decomposition (SVD) on the learned weight matrices and then analyze the behavior of VFMs after selectively removing specific \emph{singular components}\footnote{In this paper, we refer to the combination of a singular value and its corresponding singular vectors as a \emph{singular component}.}.
Our analysis reveals that singular vectors associated with higher singular values tend to extract general features spanning multiple classes in ImageNet, while minor singular components are primarily responsible for capturing context-specific features.
Capitalizing on our findings, we introduce \textbf{S}ingular Value Dec\textbf{o}mposed \textbf{M}inor Components \textbf{A}daptation (\textbf{SoMA}), which starts by performing SVD to decompose the weights into their singular vectors, \emph{i.e.,} $W=U\Sigma V^T$, then fine-tunes only the minor singular components.
Specifically, we initialize LoRA adapter using the components with the smallest $r$ singular values, \emph{i.e.,} $U_{[:,-r:]}\Sigma_{[-r:]} (V^T)_{[-r:,:]}$, and the remaining residual components are frozen to maintain generalizability, as illustrated in Fig.~\ref{fig:teaser} \emph{right}.

By extending our analysis from the weight-level to the block-level, we observe and empirically demonstrate that the early blocks of VFMs effectively extract well-localized semantic features and are less affected by style or input-level domain shifts (see Fig.~\ref{fig:early_freeze}).
Therefore, we freeze these blocks to preserve generalizable components while reducing the number of trainable parameters.
Although our proposed methods are effective, tuning the less-optimized bottom spectral space and freezing early blocks may lead to a lack of discriminability.
To mitigate this issue, we introduce an annealing weight decay scheme that gradually reduces the regularization loss incurred by weight decay over the course of training.
As shown in Fig.~\ref{fig:teaser}, comprehensive evaluations on DG benchmarks demonstrate that our framework outperforms previous state-of-the-art baselines and FFT, training only 0.58\% to 1.6\% of the parameters.
Furthermore, SoMA demonstrates impressive scaling performance with both data and model size.

\noindent We make the following contributions:
\begin{itemize}[leftmargin=3mm, itemsep=0.5mm, topsep=1mm, partopsep=0mm]
    \item We underscore the critical importance of preserving pre-trained knowledge while learning task-specific features in scenarios that involve the use of VFMs.
    \item  We present SoMA, a novel PEFT method specifically tailored for DG problem, which leverages spectral information to retain the generalization capacity of VFMs.
    In addition, we identify and freeze domain-generalizable blocks while employing an annealing weight decay strategy, thereby achieving an optimal balance between generalizability and discriminability. 
    \item Extensive experiments on various DG for semantic segmentation and object detection tasks reveal that our framework surpasses existing baselines by a substantial margin.
    Notably, SoMA achieves an mIoU of \textbf{80.4\%} on \emph{Cityscapes} without accessing real-world datasets, while training only \textbf{0.58\%} of the parameters.
    Additionally, under the most challenging \emph{daytime-sunny} $\rightarrow$ \emph{night-rainy} detection setting, SoMA outperforms the prior SOTA implementation by achieving a \textbf{+23.5\%} mAP gain.
\end{itemize}

\vspace{-2mm}
\section{Related Work}
\label{sec:related_work}
\vspace{-0.5mm}

\textbf{Domain generalization for dense predictions.} 
Domain Generalization (DG) methods for dense predictions~\cite{modify, ding2023hgformer, kim2022pin, dtp, nightseg, lee2022fifo, shen2023diga, pdod, cmformer, spc,dpcl, wu2022single-dgod, chang2024unified} have recently garnered considerable attention due to their practical demands.
These methods can be categorized along two main axes: i) data augmentation or domain randomization, both of which diversify the training process, and ii) alignment techniques to suppress domain-relevant features.

Methods adopting the first approach at the input level employ various augmentation techniques~\cite{advstyle, chattopadhyay2023pasta, peng2021globallocalTR, paintingbynumbers} or, more recently, leverage advances in generative modeling, such as diffusion models~\cite{LDM, podell2024sdxl}, to generate data approximating the target domain~\cite{jia2025dginstyle, clouds, diffusiondomainextension}, thereby expanding domain diversity and enriching learned representations.
To enable more flexible augmentation, several studies~\cite{udupa2024mrfp, NP, lee2022wildnet, famix, vidit2023clipthegap} leverage feature stylization to enhance model robustness.
On the other hand, alignment-based methods effectively reduce domain-sensitive components by employing feature normalization/whitening techniques~\cite{ibnnet, sansaw, tang2021crossnorm}, or by introducing regularization losses to suppress inconsistencies caused by simulated domain shifts~\cite{ahn2024blindnet, kim2023texture, randomization, wu2022siamdoge, shade, jing2023order, choi2021robustnet, xu2022dirl, oamix, liu2024unbiased, divalign}.
However, most of the aforementioned approaches experimented with CNN backbones~\cite{resnet} trained on relatively restricted domains. In contrast, when applying vision foundation models (VFMs)—trained on massive datasets using transformer architectures~\cite{transformer, vit} and sophisticated training schemes~\cite{oquab2023dinov2}—to the DG problem, a fundamental shift in approach is imperative.

Following this trend, recent studies leverage vision-language models as feature extractors~\cite{hummer2023vltseg}, employing text embeddings with domain-invariant semantics as a source for style augmentation~\cite{famix, vidit2023clipthegap} or as object queries in transformer-based decoders~\cite{tqdm}.
Rein~\cite{rein} injects learnable adapters 
 and tokens to refine feature maps for each instance while keeping the backbone frozen, significantly expanding the performance frontier.
FADA~\cite{bi2024fada} and SET~\cite{set} introduce frequency-adapted methods to effectively utilize frozen VFM features.
In contrast to existing methods that simply freeze all parameters, we perform singular value decomposition on the pre-trained weights, freezing the generalizable principal components and tuning only the minor singular components responsible for context-specific features, thereby achieving superior DG performance.

\noindent \textbf{Vision foundation models (VFMs).} Recently, the advent of VFMs with general perception capabilities has laid the foundation for a paradigm in which these models are broadly applicable to numerous downstream tasks.
Seminal examples of models advancing this direction include CLIP~\cite{radfordclip}, which demonstrates strong zero-shot generalizability through training on web-scale, weakly supervised image-text pairs; EVA02~\cite{eva02}, which enhances CLIP features with masked image modeling~\cite{he2022masked}; and DINOv2~\cite{oquab2023dinov2}, which incorporates losses from prior arts like iBoT~\cite{zhou2021ibot} and DINO~\cite{dino}, and is trained on massive curated datasets, offering strong spatial features for dense prediction tasks.
Our SoMA builds on this foundation, harnessing the full potential of VFMs in the context of DG problem.

\noindent \textbf{Parameter-efficient fine-tuning (PEFT).}
With the remarkable success of PEFT methods in efficiently adapting large-scale foundation models in the realm of natural language processing, there has been a growing interest in extending these approaches to the field of computer vision.
For example, VPT~\cite{jia2022vpt} prepends prompt tokens to the input sequence of several attention layers. AdaptFormer~\cite{chen2022adaptformer} introduces a lightweight adapter in parallel to the feed-forward network, whereas SSF~\cite{ssf} incorporates scale and shift parameters to modulate features for downstream tasks.
LoRA~\cite{hu2022lora} models the incremental updates of pre-trained weights using rank decomposition matrices, achieving performance comparable to full fine-tuning (FFT).
However, there still often exists a performance gap between LoRA and FFT.
DoRA~\cite{dora} and PiSSA~\cite{meng2024pissa} address this gap by further decomposing the pre-trained weights, initializing the LoRA adapter with directional and principal components, respectively, and fine-tuning accordingly.
While these methods aim to match the performance of FFT, our focus is on the preservation of generalizable components.

\vspace{-1mm}
\section{Preliminaries}
\vspace{-0.5mm}
\textbf{Low-Rank Adaptation (LoRA).}
LoRA~\cite{hu2022lora} hypothesizes that the weight changes during fine-tuning exhibit a low-rank structure.
Given a pre-trained weight matrix $W_0 \in \mathbb{R}^{m \times n}$, LoRA constrains its update $\Delta W \in \mathbb{R}^{m \times n}$ to a low-rank decomposition $\Delta W = BA$, where $B \in \mathbb{R}^{m \times r}$ and $A \in \mathbb{R}^{r \times n}$ are two low-rank matrices, with an intrinsic rank of $r \ll \text{min}(m, n)$.
For $y = W_0x$, the modified forward pass is as follows:
\vspace{-1.6mm}
\begin{equation}
    y = (W_0 + \Delta W)x  = (W_0+\underline{BA})x,
    \label{equ:lora}
    \vspace{-1.5mm}
\end{equation}
where $W_0$ is frozen, and only the underlined low-rank parameters are fine-tuned, significantly reducing the number of trainable parameters.
LoRA uses a uniform Kaiming distribution~\cite{heinit} to initialize $A$, while $B$ is initially set to zero, ensuring $BA=0$ at the beginning of training.
As seen in Eq.~\ref{equ:lora}, the learned matrices $BA$ can be merged with the frozen weight $W_0$, allowing LoRA-based variants to avoid adding any additional inference burden.

\noindent \textbf{Singular Value Decomposition (SVD) Analysis.}
We observe that LoRA and existing PEFT methods, by training only a relatively small subset of parameters or by enforcing low-rank-ness of weight updates, deliver promising DG performance without significantly distorting the pre-trained representations.
However, many previous studies build adapters without considering how the generalizable components are structured within the weight matrices, potentially interfering with these components in the fine-tuning process.
Building on the Eckart–Young–Mirsky theorem~\cite{svd}, which validates that the optimal rank-$r$ approximation of a matrix $W$ is represented by the sum of its top $r$ singular components, we analyze how the world knowledge of VFMs is structured from an SVD perspective.
Using ImageNet-1k~\cite{deng2009imagenet} as a testbed, our analysis computes the SVD of pre-trained weights across all layers of the DINOv2~\cite{oquab2023dinov2} ViT-large model, followed by truncating specific singular values and their corresponding singular vectors to investigate the classes that are subsequently misclassified.

\begin{figure}[t]
    \vspace{-3.4mm}
    \centering
    \includegraphics[width=0.8\linewidth, keepaspectratio]{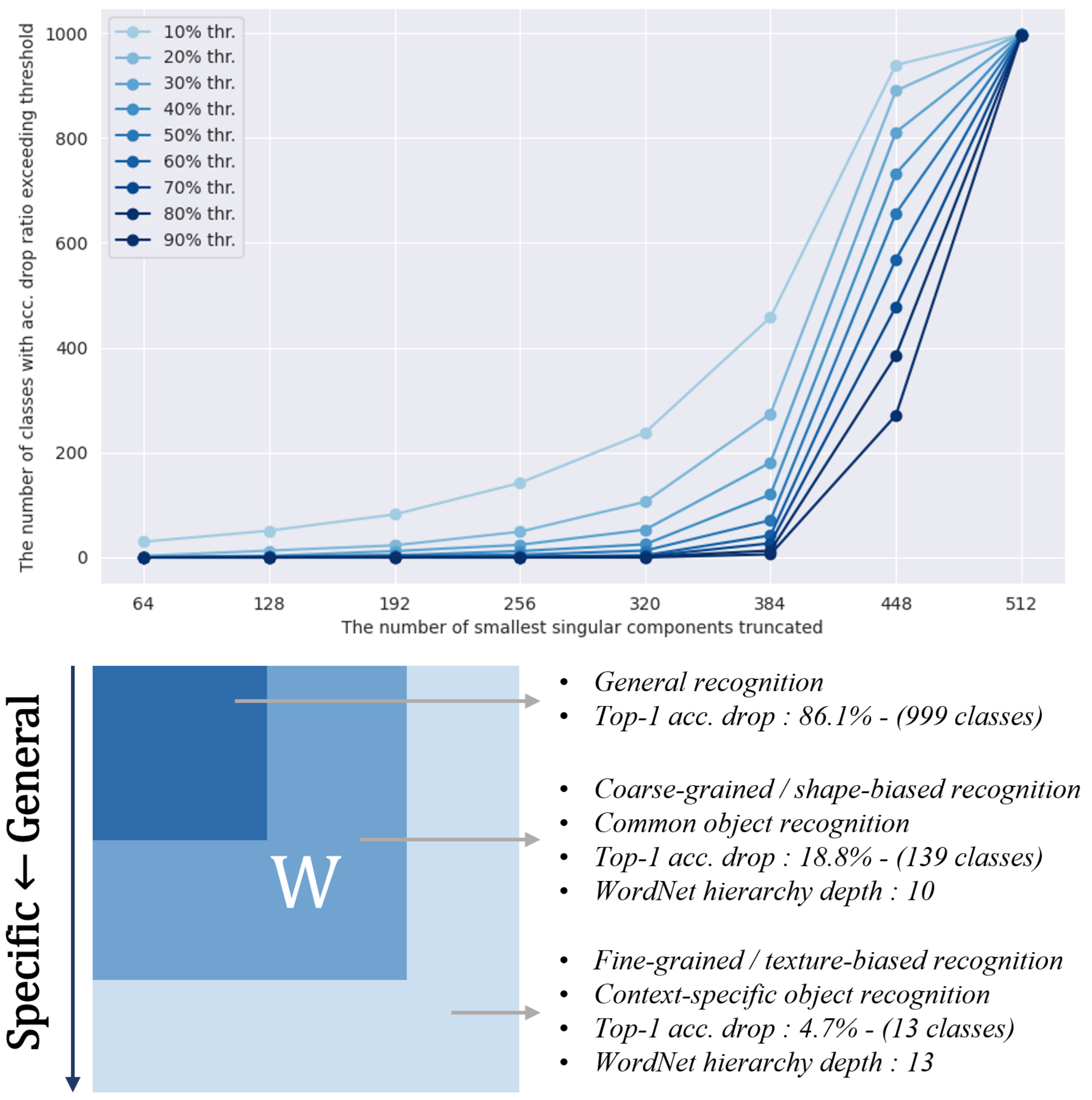}
    \vspace{-3mm}
    \caption{\textbf{Distribution of generalizable components.}
    \emph{Top.} Number of classes exhibiting specific accuracy drops after applying SVD to DINOv2-large weights and reconstructing by truncating the smallest $r$ singular components.
    \emph{Bottom.} Distinct roles of singular components across levels. Numbers in parentheses represent the count of classes with an accuracy drop ratio exceeding 50\%.
    The average WordNet hierarchy depth of these classes is shown, with higher values indicating greater context specificity.
    }
    \label{fig:rank_reduction}
    \vspace{-5.6mm}
\end{figure}

As shown in Fig.~\ref{fig:rank_reduction} \emph{top}, as the number of discarded smallest singular components increases, the number of misclassified classes grows exponentially rather than linearly, suggesting that components with larger singular values tend to capture more general features spanning multiple classes.
To further examine whether different groups of singular components exhibit distinct behaviors, we consider three groups: the top 8 principal components $U_{[:,:8]}\Sigma_{[:8]} (V^T)_{[:8,:]}$, the middle 160 singular components $U_{[:,384:544]}\Sigma_{[384:544]} (V^T)_{[384:544,:]}$, and the bottom 320 minor components $U_{[:,-320:]}\Sigma_{[-320:]} (V^T)_{[-320:,:]}$.
We then truncate each group separately and analyze the classes that the rank-reduced model subsequently fails to classify (see Fig.~\ref{fig:rank_reduction} \emph{bottom}).

According to the Eckart–Young–Mirsky theorem and the heavy-tail distribution of singular values, it is evident that the top singular components encapsulate the core knowledge of VFM and should be preserved to maintain generalizability.
Conversely, we note that the bottom-component-truncated model  fails to recognize classes that require fine-grained or texture-biased features (e.g., specific dog breeds, wildlife with distinctive textures), as well as those that are context-specific (e.g., cornet, bikini, missile, dock).
In contrast, we find that removing middle singular components results in significant performance declines for classes that rely on coarse-grained or shape-biased recognition (e.g., mortarboard, pier, altar, necklace, schooner, horizontal bar).
Interestingly, as suggested by differences in the depth of the WordNet hierarchy, higher-value singular components do not function independently but rather interact in a hierarchical and composite manner.
For instance, classes with notable accuracy drops can also be clustered into higher-level concepts, such as \emph{functionality} (e.g., wok, hotpot, pitcher, cauldron, pop bottle, cup/ambulance, minivan, passenger car, police van, garbage truck, limousine) or \emph{co-occurrence} (e.g., desktop computer, keyboard, notebook, desk).
Additionally, almost all of these objects are commonly found in everyday scenes.
\emph{Consequently, it can be seen that generalizable components are hierarchically and compositionally structured according to the magnitude of singular values.}

\noindent\textbf{Problem Formulation.}
The entire model consists of a feature extractor $\mathcal{F}$, which combines a VFM parameterized by $\Phi$ with our SoMA adapter parameterized by $\theta_{S}$, and a task-specific decode head $\mathcal{H}$, parameterized by $\theta_{h}$.
The optimization objective can be expressed as follows:
\vspace{-1.5mm}
\begin{equation}
\label{eq:dg objective}
\mathop{\arg\min}\limits_{\theta_{S},\theta_{h}} 
\sum_{i=1}^{N_s} \mathcal{L}_{task}(\mathcal{H}_{\theta_{h}}(\mathcal{F}_{\Phi,\theta_{S}}(x_i)),y_i),
\vspace{-1mm}
\end{equation}
where ($x_i, y_i$) denotes an input image and its associated ground truth, drawn from the source domain datasets $\mathcal{D}_{s}$, comprising $N_s$ samples.
Our proposed SoMA framework facilitates generalizing task-specific skills learned from $\mathcal{D}_{s}$ to the unseen target domain datasets $\mathcal{D}_{t}$.

\vspace{-1.5mm}
\section{Proposed Method}
\vspace{-1.3mm}
In this section, we present our parameter-efficient adaptation methods specifically tailored for DG dense prediction tasks, inspired by the insights from our SVD analysis. Subsequently, we propose a block freeze strategy and an annealing weight decay scheme to strike an optimal trade-off between generalizability and discriminability.

\vspace{-1mm}
\subsection{Minor Singular Components Adaptation}
To effectively \textbf{preserve} the integrity of generalizable representations in pre-trained weights, we propose Singular value decomposed Minor components Adaptation (SoMA).
Firstly, SoMA performs singular value decomposition (SVD) on the pre-trained weight matrices, including those from the attention layers, multi-layer perceptron (MLP) layers, and more.
The weight matrix $W \in \mathbb{R}^{m \times n}$ is decomposed using SVD as follows:
\vspace{-2.5mm}
\begin{equation}
    W = U\Sigma V^T=\sum_{i=1}^{R} \sigma_i u_i v_i^T,
    \label{equ:svd}
\end{equation}

\vspace{-1.5mm}
\noindent where $U = [u_1, u_2, \cdots, u_m] \in \mathbb{R}^{m \times m}$ and $V = [v_1, v_2, \cdots, v_n]  \in \mathbb{R}^{n \times n}$ are the left and right singular vectors, whose columns form an orthonormal basis for  $\mathbb{R}^m$ and $\mathbb{R}^n$ respectively, $\Sigma \in \mathbb{R}^{m \times n}$ is a diagonal matrix whose entries $\sigma_i$ are singular values arranged in descending order, and $R$ denotes the rank of $W$, with $R\le \min(m,n)$.

We then divide the singular components into two groups according to their singular values: the minor singular components with the smallest $r$ singular values $U_{[:,-r:]}\Sigma_{[-r:]} (V^T)_{[-r:,:]}$ and the residual singular components $U_{[:,:-r]}\Sigma_{[:-r]} (V^T)_{[:-r,:]}$.
As illustrated in Fig.~\ref{fig:teaser} \emph{right}, we employ the QR-type reconstruction of minor singular components to initialize the adapter for fine-tuning:
\vspace{-2mm}
\begin{align}
B &= U_{[:,-r:]} \sqrt{\Sigma}_{[-r:]} \in \mathbb{R}^{m \times r},\label{equ:init_b}\\
A &= \sqrt{\Sigma}_{[-r:]} (V^T)_{[-r:,:]} \in \mathbb{R}^{r \times n}.\label{equ:init_a}
\vspace{-2mm}
\end{align}
\begin{figure}[t]
    \vspace{-3.4mm}
    \centering
    \includegraphics[width=\linewidth, height=5.5cm]{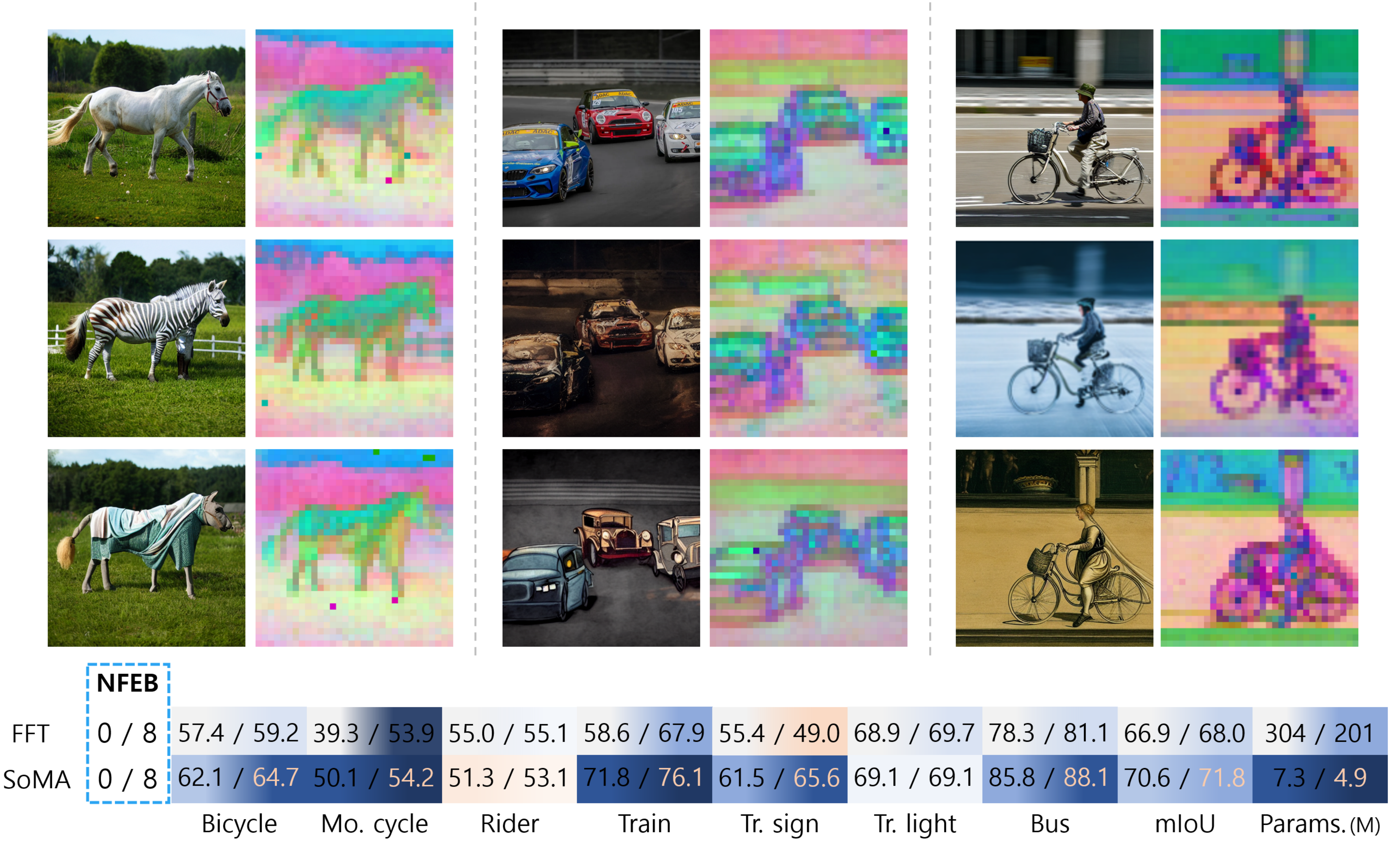}
    \vspace{-6mm}
    \caption{\textbf{The inherent generalization capabilities of the early blocks of VFM.} 
    \emph{Top.} We apply PCA on the extracted intermediate features (8th block of DINOv2-Large) and visualize the top three leading components.
    \emph{Bottom.} Class-wise IoU comparison for rare classes~\cite{hoyer2022daformer} under the \emph{GTAV}$\rightarrow$\emph{Cityscapes} setting, focusing on the impact of freezing the first eight blocks. “NFEB” stands for the Number of Frozen Early Blocks. A darker shade of blue indicates enhanced generalization performance relative to the baseline.
    }
    \label{fig:early_freeze}
    \vspace{-5mm}
\end{figure}
In the fine-tuning process, the frozen residual matrix $W_{res}$ is defined as $W - BA$ to prevent numerical errors introduced during the SVD step, while maintaining the pre-trained generalization capability at the beginning of fine-tuning.
After fine-tuning, the learned low-rank matrices $B'$ and $A'$ can be merged into $W_{res}$, as in LoRA~\cite{hu2022lora}, ensuring no additional computation overhead during inference:
\vspace{-1.3mm}
\begin{equation}
    y = (W \underbrace{-BA +B'A'}_{\Delta W_{\scriptscriptstyle SoMA}})x  = (W_{res}+B'A')x = W'x.
    \label{equ:merge}
    \vspace{-1.5mm}
\end{equation}

Importantly, SoMA offers several key advantages for tackling the DG problem: i) SoMA initializes the adapter using components orthogonal to the principal components, thus minimizing interference with pre-trained representations.
For the batched version of the forward pass described in Eq.~\ref{equ:lora} (with $X \in \mathbb{R}^{n \times b}$ and $Y \in \mathbb{R}^{m \times b}$), the gradients of $B$ and $A$ are $\frac{\partial \mathcal{L}}{\partial B} = \frac{\partial \mathcal{L}}{\partial Y} X^T A^T$ and $\frac{\partial \mathcal{L}}{\partial A} = B^T \frac{\partial \mathcal{L}}{\partial Y} X^T$.
Therefore, unlike LoRA, which randomly initializes $A$, our approach facilitates convergence along singular directions that preserve the integrity of the generalizable components.
ii) Consistent with prior DG methods~\cite{choi2021robustnet, lee2022wildnet, kim2023texture, advstyle} that guide models to prioritize shape over texture, thereby achieving superior out-of-distribution performance, SoMA effectively preserves the domain generalizable shape-biased features of VFMs by tuning only the minor singular components (see Fig.~\ref{fig:rank_reduction} \emph{bottom} and Tab.~\ref{tab: smr}).

\begin{table}[t]
    \begin{minipage}{0.48\textwidth}
    \centering
        \vspace{-3.4mm}
        \tablestyle{3.2pt}{0.95}
        \begin{tabular}{c|c|c|ccc}\toprule
SMR & Block & $i$ & $\Delta W_{\scriptscriptstyle SoMA}$ & $\Delta W_{\scriptscriptstyle SoMA}^*$ & $\Delta W_{\scriptscriptstyle LoRA}$\\
\midrule
\multirow{8}{*}{$|\frac{u_i^\mathsf{T}\Delta Wv_i}{\sigma_i}|$} & \multirow{4}{*}{12th}     
   & 0–255 & 0.075 & 0.084 {\tiny \textcolor{red}{($\downarrow$ 17\%)}} & 0.101 \\
 & & 256–511 & 0.094 & 0.094 {\tiny \textcolor{red}{($\downarrow$ 43\%)}}& 0.166 \\
 & & 512–767 & 0.157 & 0.165 {\tiny \textcolor{red}{($\downarrow$ 40\%)}}& 0.275 \\
 & & 768–1023 & 4.104& 6.152 {\tiny \textcolor{blue}{($\uparrow$ 21\%)}} & 5.095 \\
 \cmidrule(r){2-6}
 & \multirow{4}{*}{24th} & 0–255 & 0.097 & 0.095 {\tiny \textcolor{red}{($\downarrow$ 24\%)}}& 0.125 \\
 & & 256–511 & 0.105 & 0.108 {\tiny \textcolor{red}{($\downarrow$ 35\%)}}& 0.166 \\
 & & 512–767 & 0.206 & 0.189 {\tiny \textcolor{red}{($\downarrow$ 41\%)}} & 0.323 \\
 & & 768–1023 & 2.781 & 3.620 {\tiny \textcolor{blue}{($\uparrow$ 26\%)}} & 2.865 \\
\bottomrule
\end{tabular}

        \vspace{-2.2mm}
        \caption{\textbf{Singular modulation ratio comparison.}
        * indicates the adaptation matrix trained using the annealing weight decay strategy. The weight matrices are taken from the query projection layers of the 12th and 24th attention blocks in DINOv2-Large~\cite{oquab2023dinov2}.
        }\label{tab: smr}
        \vspace{-5mm}
    \end{minipage}
\end{table}

\noindent \textbf{Discussion.}
To assess whether the proposed adaptation matrix $\Delta W_{\scriptscriptstyle SoMA}$ minimally correlates with the top singular directions of the pre-trained weight matrix $W_0$, we project $\Delta W$ onto the singular vectors of $W_0$.
Specifically, we quantify the extent of interference by $\Delta W$ on each $i$-th singular direction through a \emph{singular modulation ratio} (SMR), which is formally defined as follows:
\vspace{-2mm}
\begin{equation}
    \mathrm{SMR}_i = |\frac{u_i^\mathsf{T}\Delta Wv_i}{\sigma_i}|.
    \label{equ:smr}
    \vspace{-1.5mm}
\end{equation}
For ease of analysis, we partition the 1024 singular directions into four equal groups, calculate the average SMR for each group, and then compare the resulting values between SoMA and LoRA.
As shown in Tab.~\ref{tab: smr}, the SoMA initialization scheme proves more effective than LoRA in minimizing interference with singular components with higher singular values.
In summary, our method effectively preserves the structure of the VFM's generalizable representations while adapting to downstream tasks, resulting in significant improvements on DG benchmarks compared to SOTA baselines, as confirmed by our experimental results.

\vspace{-1.5mm}
\subsection{Freezing Early Blocks}
\vspace{-0.5mm}
In this section, we explore the distribution of generalizable components at the block level.
Drawing on recent domain adaptation methods~\cite{lee2023surgical, dplot}, which find that the early blocks of a model primarily engage in domain-specific feature extraction and should therefore be updated, we conjecture that in the context of leveraging VFM for DG, the early blocks should instead remain unchanged.
Intuitively, these initial blocks of VFMs are adept at projecting low-level features from diverse domains into domain-invariant semantic representations.
In other words, while \textbf{adjusting early blocks} may help bridge input-level domain gaps in domain adaptation, \textbf{freezing the early blocks of VFMs} is essential for DG to prevent overfitting to the source domain and to retain the model's generalization capability.

To validate our hypothesis,  we perform PCA analysis on the features extracted from the early blocks of VFM and compare the performance on rare classes when these blocks are frozen, as illustrated in Fig.~\ref{fig:early_freeze}.
Despite considerable domain shifts in style and content, the principal components align well with the image layout and maintain consistent semantic information (e.g., the horses have similar colors).
Furthermore, simply freezing the early blocks yields a notable improvement in IoU for rare classes, indicating that updating these blocks can severely disrupt the learned representations of various classes in VFMs.
Thus, by default, we opt to freeze all blocks up to the one that generates the feature map fed as the highest resolution input to the segmentation/detection head (e.g., the first eight blocks).
Detailed ablations of varying the number of frozen early blocks are presented in the \emph{Supplemental}.

\vspace{-1mm}
\subsection{Annealing Weight Decay}
Our methods are specialized in preserving the pre-trained knowledge of VFMs, but this may come at the cost of diminished discriminability.
One can remove weight decay or set its coefficient to a very small value to enhance discriminability.
However, doing so risks increased interference of the tuned parameters with the principal components of the pre-trained weights.
Therefore, based on the observation that regularization primarily affects the early phases of learning by guiding a model towards having good generalization properties~\cite{time_matters}, we leverage an Annealing Weight Decay (AWD) scheme.
Specifically, AWD starts with a relatively large weight decay coefficient and progressively reduces it to zero as training progresses.
By default, we use a cosine schedule for AWD.
As shown in Tab.~\ref{tab: smr} ($\Delta W_{\scriptscriptstyle SoMA}$ \emph{vs.} $\Delta W_{\scriptscriptstyle SoMA}^*$) and Tab.~\ref{tab: component_ablation}, AWD integrates seamlessly with SoMA, improving discriminability \emph{without compromising SoMA's ability to preserve pre-trained knowledge}.

\vspace{-2mm}
\section{Experiments}
\vspace{-1mm}
We conduct a variety of experiments to showcase the efficacy of SoMA on DG benchmarks including domain generalized semantic segmentation (DGSS) and domain generalized object detection (DGOD).
We then perform an ablation study on SoMA, emphasizing the contribution of each component to the DG performance.
Lastly, we extend our approach beyond dense predictions to generative modeling.

\vspace{-1mm}
\subsection{Experimental Setup}
\vspace{-0.6mm}
\textbf{Datasets.} \underline{For DGSS}, we use three synthetic datasets (GTAV~\cite{gtav}, SYNTHIA~\cite{synthia}, UrbanSyn~\cite{urbansyn}), and four real-world datasets (Cityscapes~\cite{citys}, BDD100K~\cite{bdd}, Mapillary~\cite{map}, ACDC~\cite{acdc}).
Specifically, GTAV, a large-scale dataset generated from the game engine, comprises 24,966 images.
SYNTHIA consists of 9,400 photo-realistic synthetic images.
UrbanSyn provides 7,539 driving scene images.
Cityscapes is an autonomous driving dataset containing 2,975 training and 500 validation images (2048$\times$1024).
BDD and Mapillary have 1,000 (1280$\times$720) and 2,000 (1920$\times$1080) validation images, respectively.
ACDC is a driving scene dataset under adverse conditions, including night, snow, fog, and rain.
\underline{For DGOD}, we use the urban scene detection dataset introduced by \cite{wu2022single-dgod}.
\begin{table}[t]
    \begin{minipage}{0.48\textwidth}
    \centering
        \tablestyle{1.0pt}{1.01}
        \begin{tabular}{lcccccc} 
\toprule
\multicolumn{3}{c}{\textbf{\emph{Synthetic-to-Real Generalization}}}&\multicolumn{4}{c}{Test Domains (mIoU in \%)} \\ \cmidrule(r){4-7}
Methods &Backbone &$\text{Params.}^{*}$ &$\rightarrow$\emph{Citys.} &$\rightarrow$\emph{BDD} &$\rightarrow$\emph{Map.} &Avg. \\\midrule
\multicolumn{7}{c}{\emph{Single-source DGSS Trained on \underline{GTAV}}} \\
\addlinespace[2pt]
\normal CLOUDS~\cite{clouds} &CLIP-CN-L &0.0M &60.20 &57.40 &67.00 &61.50 \\
\normal VLTSeg~\cite{hummer2023vltseg} &EVA02-L &304.2M &65.30 &58.30 &66.00 &63.20 \\
\normal Rein~\cite{rein} &EVA02-L &3.0M &65.30 &60.50 &64.90 &63.60 \\
\normal FADA~\cite{bi2024fada} &EVA02-L &11.7M &66.70 &61.90 &66.10 &64.90 \\ 
\normal tqdm~\cite{tqdm} &EVA02-L &304.2M &68.88 &59.18 &70.10 &66.05 \\
\cellcolor[HTML]{efefef}\normal SoMA~(Ours) &\cellcolor[HTML]{efefef}EVA02-L & \cellcolor[HTML]{efefef}5.1M &\cellcolor[HTML]{efefef}68.05 &\cellcolor[HTML]{efefef}60.81 &\cellcolor[HTML]{efefef}68.33 &\cellcolor[HTML]{efefef}65.73 \\
\cellcolor[HTML]{efefef}\tta SoMA~(Ours) &\cellcolor[HTML]{efefef}EVA02-L &\cellcolor[HTML]{efefef}5.1M&\cellcolor[HTML]{efefef}69.94 &\cellcolor[HTML]{efefef}62.48&\cellcolor[HTML]{efefef}68.33 &\cellcolor[HTML]{efefef}66.92 \\
\peft DoRA~\cite{dora} &DINOv2-L &7.5M & 66.12 &59.31 &67.07 & 64.17 \\
\peft VPT~\cite{jia2022vpt} &DINOv2-L &3.7M & 68.75 &58.64 &68.32 & 65.24 \\
\normal SET~\cite{set} &DINOv2-L &6.1M &68.06 &61.64 &67.68 &65.79 \\
\normal FADA~\cite{bi2024fada} &DINOv2-L &11.7M &68.23 &61.94 &68.09 & 66.09 \\
\peft \scalebox{0.85}{AdaptFormer}~\cite{chen2022adaptformer} &DINOv2-L &6.3M &70.10 & 59.81 & 68.77 & 66.23 \\
\peft SSF~\cite{ssf} &DINOv2-L &0.5M & 68.97 &61.30 & 68.77 &66.35 \\
\peft LoRA~\cite{hu2022lora} &DINOv2-L &7.3M & 70.13 &60.13 &70.42 & 66.89 \\
\tta $\text{Rein}^{\dagger}$~\cite{rein} &DINOv2-L &3.0M &70.68 &62.51 &69.61 &67.60 \\
\cellcolor[HTML]{efefef}\normal SoMA~(Ours) &\cellcolor[HTML]{efefef}DINOv2-L & \cellcolor[HTML]{efefef}4.9M &\cellcolor[HTML]{efefef}71.82 &\cellcolor[HTML]{efefef}61.31 &\cellcolor[HTML]{efefef}\textbf{71.67} &\cellcolor[HTML]{efefef}68.27 \\
\cellcolor[HTML]{efefef}\tta SoMA~(Ours) &\cellcolor[HTML]{efefef}DINOv2-L &\cellcolor[HTML]{efefef}4.9M&\cellcolor[HTML]{efefef}\textbf{73.63} &\cellcolor[HTML]{efefef}\textbf{63.33}&\cellcolor[HTML]{efefef}70.98 &\cellcolor[HTML]{efefef}\textbf{69.31} \\
\midrule
\multicolumn{7}{c}{\emph{Multi-source DGSS Trained on \underline{GTAV + SYNTHIA}}} \\
\addlinespace[2pt]
\normal $\text{Rein}^{\dagger}$~\cite{rein} &DINOv2-L &3.0M &72.17 &61.53 &70.69 &68.13 \\
\cellcolor[HTML]{efefef}\normal SoMA~(Ours) &\cellcolor[HTML]{efefef}DINOv2-L & \cellcolor[HTML]{efefef}4.9M &\cellcolor[HTML]{efefef}73.16 &\cellcolor[HTML]{efefef}61.90 &\cellcolor[HTML]{efefef}72.73 &\cellcolor[HTML]{efefef}69.26 \\
\cellcolor[HTML]{efefef}\tta SoMA~(Ours) &\cellcolor[HTML]{efefef}DINOv2-L &\cellcolor[HTML]{efefef}4.9M&\cellcolor[HTML]{efefef}\textbf{74.85} &\cellcolor[HTML]{efefef}\textbf{63.59}&\cellcolor[HTML]{efefef}\textbf{73.92} &\cellcolor[HTML]{efefef}\textbf{70.79} \\
\midrule
\multicolumn{7}{c}{\emph{Multi-source DGSS Trained on \underline{GTAV + SYNTHIA + UrbanSyn}}} \\
\addlinespace[2pt]
\normal FFT &DINOv2-L &304.2M &75.90 &60.93 &72.80 &69.88 \\
\cellcolor[HTML]{efefef}\normal SoMA~(Ours) &\cellcolor[HTML]{efefef}DINOv2-L & \cellcolor[HTML]{efefef}4.9M &\cellcolor[HTML]{efefef}77.33 &\cellcolor[HTML]{efefef}62.78 &\cellcolor[HTML]{efefef}74.93 &\cellcolor[HTML]{efefef}71.68 \\
\normal $\text{FFT}^{\ddagger}$ &DINOv2-L &307.3M &77.06 &61.81 &75.09 &71.32 \\
\normal $\text{Rein}^{\ddagger}$~\cite{rein} &DINOv2-L &3.0M &78.42 & 62.20 &74.49 &71.70 \\
\cellcolor[HTML]{efefef}\normal $\text{SoMA}^{\ddagger}$ (Ours) &\cellcolor[HTML]{efefef}DINOv2-L & \cellcolor[HTML]{efefef}4.9M &\cellcolor[HTML]{efefef}79.22 &\cellcolor[HTML]{efefef}63.84 &\cellcolor[HTML]{efefef}\textbf{76.30} &\cellcolor[HTML]{efefef}73.12 \\
\normal Freeze &DINOv2-G & 0.0M &76.08 &61.98 &72.23 &70.10 \\
\normal FFT &DINOv2-G & 1.1B &76.90 &61.69 &73.53 &70.71 \\
\cellcolor[HTML]{efefef}\normal SoMA~(Ours) &\cellcolor[HTML]{efefef}DINOv2-G & \cellcolor[HTML]{efefef}6.6M &\cellcolor[HTML]{efefef}78.39 &\cellcolor[HTML]{efefef}63.75 &\cellcolor[HTML]{efefef}75.16 &\cellcolor[HTML]{efefef}72.43 \\
\cellcolor[HTML]{efefef}\tta SoMA~(Ours) &\cellcolor[HTML]{efefef}DINOv2-G &\cellcolor[HTML]{efefef}6.6M &\cellcolor[HTML]{efefef}\textbf{80.37} &\cellcolor[HTML]{efefef}\textbf{65.67}&\cellcolor[HTML]{efefef}76.18 &\cellcolor[HTML]{efefef}\textbf{74.07} \\
\bottomrule
\end{tabular}
        \vspace{-3mm}
        \caption{
        Comparison of the proposed SoMA with existing DGSS \normal and PEFT \peft methods under various \textbf{synthetic-to-real settings}.
        }\label{tab: dgss_s2r}
        \vspace{-6mm}
    \end{minipage}
\end{table}
The dataset consists of five different weather conditions: Daytime-Sunny (DS), Night-Clear (NC), Night-Rainy (NR), Dusk-Rainy (DR), and Daytime-Foggy (DF).
Daytime-Sunny serves as the source domain, offering 26,518 images, with 8,313 used for in-domain evaluation, while the remaining adverse weather conditions are employed as unseen target domains.

\noindent \textbf{Implementation details.} Since our method focuses on preserving pre-trained knowledge, we mainly evaluate its effectiveness using VFMs like EVA02~\cite{eva02} and DINOv2~\cite{oquab2023dinov2} as the backbone.
To further elevate performance, we leverage state-of-the-art decode heads such as Mask2Former~\cite{cheng2022mask2former} and Co-DETR~\cite{codetr}.
Extensive experiments with various backbones and heads are provided in the \emph{Supplemental}.
By default, SoMA is applied to all linear layers within the self-attention and MLP, with a rank of 16.
Detailed training settings are presented in the \emph{Supplemental}. 

\vspace{-1mm}
\subsection{Comparison with State-of-the-Arts}
\vspace{-0.5mm}
We compare SoMA with previous SOTA baselines, achieving the highest performance across all DG scenarios \emph{without adding inference latency or regularization loss}.
In our tables, $*$ indicates trainable parameters in the backbones.
Marker \tta refers to models tested with multi-scale flip augmentation.
$\dagger$ represents re-implemented test results using official checkpoints for a fair comparison, and $\ddagger$ denotes training on images cropped to 1024$\times$1024; otherwise, the default is 512$\times$512.
A comprehensive comparison with earlier works can be found in the \emph{Supplemental}.
\begin{table}[t]
    \begin{minipage}{0.48\textwidth}
    \centering
        \tablestyle{3.2pt}{1.0}
        \begin{tabular}{lccccc}  
\toprule
\multicolumn{3}{c}{\textbf{\emph{Real-to-Real Generalization}}}&\multicolumn{3}{c}{Test Domains (mIoU in \%)} \\ \cmidrule(r){4-6}
Methods &Backbone &$\text{Params.}^{*}$ &$\rightarrow$\emph{BDD} &$\rightarrow$\emph{Map.} &Avg. \\\midrule
\multicolumn{6}{c}{\emph{Single-source DGSS Trained on \underline{Cityscapes}}} \\
\addlinespace[2pt]
\normal HGFormer~\cite{ding2023hgformer} &Swin-L &196.0M &61.50 &72.10 &66.80   \\
\normal CMFormer~\cite{cmformer} &Swin-L &196.0M &62.60 &73.60 &68.10 \\
\normal tqdm~\cite{tqdm} &EVA02-L &304.2M &64.72 &76.15 &70.44 \\
\normal FADA~\cite{bi2024fada} &DINOv2-L &11.7M &65.12 &75.86 &70.49 \\
\normal $\text{Rein}^{\dagger}$~\cite{rein} &DINOv2-L &3.0M &66.53 &75.18 &70.86 \\
\cellcolor[HTML]{efefef}\normal SoMA~(Ours) &\cellcolor[HTML]{efefef}DINOv2-L & \cellcolor[HTML]{efefef}4.9M &\cellcolor[HTML]{efefef}67.02 &\cellcolor[HTML]{efefef}76.45 &\cellcolor[HTML]{efefef}71.74 \\
\cellcolor[HTML]{efefef}\tta SoMA~(Ours) &\cellcolor[HTML]{efefef}DINOv2-L &\cellcolor[HTML]{efefef}4.9M&\cellcolor[HTML]{efefef}\textbf{68.08} &\cellcolor[HTML]{efefef}\textbf{77.87}&\cellcolor[HTML]{efefef}\textbf{72.98} \\
\bottomrule
\end{tabular}
        \vspace{-3mm}
        \caption{
        \textbf{Real-to-real DGSS comparison.}
        }\label{tab: dgss_r2r}
        \vspace{1mm}
    \end{minipage} \\
    \begin{minipage}{0.48\textwidth}
    \centering
        \tablestyle{4.2pt}{1.0}
        \begin{tabular}{lccccc}  
\toprule
\textbf{\emph{\scalebox{0.85}{Clear-to-Adverse Weather}}} &\multicolumn{5}{c}{ACDC~\cite{acdc} Test Domains (mIoU in \%)} \\ \cmidrule(lr){2-6}
Methods &$\rightarrow$\emph{Night} &$\rightarrow$\emph{Snow} &$\rightarrow$\emph{Fog} &$\rightarrow$\emph{Rain} &All \\\midrule
\multicolumn{6}{c}{\emph{Single-source DGSS Trained on \underline{Cityscapes}}} \\
\addlinespace[2pt]
\normal HGFormer~\cite{ding2023hgformer} &52.7 &68.6 &69.9 &72.0 &67.2   \\
\cellcolor[HTML]{efefef}\normal SoMA (Ours) &\cellcolor[HTML]{efefef}61.7 &\cellcolor[HTML]{efefef}77.3 &\cellcolor[HTML]{efefef}74.7 &\cellcolor[HTML]{efefef}77.8 &\cellcolor[HTML]{efefef}74.4 \\
\normal $\text{VLTSeg}^{\ddagger}$~\cite{hummer2023vltseg} &- &- &- &- &77.9 \\
\normal $\text{Rein}^{\ddagger}$~\cite{rein} &70.6 &79.5 &76.4 &79.4 &77.6 \\
\normal $\text{FFT}^{\ddagger}$ &68.9 &\textbf{80.3} &\textbf{76.9} &79.7 &77.7 \\
\cellcolor[HTML]{efefef}\normal $\text{SoMA}^{\ddagger}$ (Ours) &\cellcolor[HTML]{efefef}\textbf{73.2} & \cellcolor[HTML]{efefef}79.8 &\cellcolor[HTML]{efefef}76.8 &\cellcolor[HTML]{efefef}\textbf{80.2} &\cellcolor[HTML]{efefef}\textbf{78.8} \\
\bottomrule
\end{tabular}
        \vspace{-3mm}
        \caption{
        Results on \textbf{Cityscapes $\rightarrow$ ACDC \underline{test} set.}
        }\label{tab: dgss_acdc}
        \vspace{1mm}
    \end{minipage} \\
    \begin{minipage}{0.48\textwidth}
    \centering
        \tablestyle{1.5pt}{1.0}
        \begin{tabular}{lcccccc} 
\toprule
\multicolumn{3}{c}{\textbf{\emph{Data Efficiency}}}&\multicolumn{4}{c}{Test Domains (mIoU in \%)} \\ \cmidrule(r){4-7}
Methods &Backbone &$\text{Params.}^{*}$ &$\rightarrow$\emph{Citys.} &$\rightarrow$\emph{BDD} &$\rightarrow$\emph{Map.} &Avg. \\\midrule
\multicolumn{7}{c}{\emph{\scalebox{0.9}{DGSS Pre-trained on GTAV + SYNTHIA + UrbanSyn} $\rightarrow$ \underline{$\frac{1}{16}$ of Cityscapes}}} \\
\addlinespace[3.5pt]
\normal $\text{FFT}^{\ddagger}$ &DINOv2-L &307.3M &81.53 &65.22 &75.73 &74.16 \\
\normal $\text{Rein}^{\ddagger}$~\cite{rein} &DINOv2-L &3.0M &\textbf{82.58} &64.76 &73.73 &73.69\\
\cellcolor[HTML]{efefef}\normal $\text{SoMA}^{\ddagger}$ (Ours) &\cellcolor[HTML]{efefef}DINOv2-L &\cellcolor[HTML]{efefef}4.9M &\cellcolor[HTML]{efefef}82.50 & \cellcolor[HTML]{efefef}\textbf{66.99} &\cellcolor[HTML]{efefef}\textbf{77.02} &\cellcolor[HTML]{efefef}\textbf{75.50} \\
\bottomrule
\end{tabular}
        \vspace{-3mm}
        \caption{
        \textbf{Real-world data efficiency.}
        }\label{tab: dgss_data_efficiency}
        \vspace{-6.8mm}
    \end{minipage}
\end{table}

\noindent \textbf{Synthetic-to-real DGSS.} In Tab.~\ref{tab: dgss_s2r}, we adopt \emph{GTAV} $\rightarrow$ \{\emph{Citys.}, \emph{BDD}, \emph{Map.}\} as the basic setting, and demonstrate scalability of SoMA by incorporating additional synthetic datasets into the training process.
We report the mean Intersection over Union (mIoU) and benchmark our results against SOTA methods.
SoMA consistently outperforms VFM-based DGSS methods and PEFT approaches using both vision-language (EVA02) and vision-only (DINOv2) VFM backbones.
Notably, SoMA achieves superior results over LoRA by minimally interfering with the hierarchical structure of VFM's pre-trained knowledge.
Moreover, SoMA exhibits superior scalability across data, model, and input size, with improvements becoming more pronounced when test-time augmentation is applied.
We provide comparisons regarding model efficiency in the \emph{Supplemental}.

\noindent \textbf{Real-to-real DGSS.} In Tables~\ref{tab: dgss_r2r} and~\ref{tab: dgss_acdc}, we conduct experiments under the \emph{Citys.} $\rightarrow$ \{\emph{BDD}, \emph{Map.}\}\scalebox{1.1}{/}\emph{ACDC} settings.
When combined with DINOv2-L backbone, SoMA consistently gains the best results across all settings.
This highlights the effectiveness of our method as a strong solution for real-world deployment scenarios, where domain shifts caused by geographic or weather variations are common.
\begin{table}[t]
    \begin{minipage}{0.48\textwidth}
    \centering
        \tablestyle{2.4pt}{1.0}
        \begin{tabular}{lccccccc}  
\toprule
\multicolumn{3}{c}{\textbf{\emph{\scalebox{0.95}{Clear-to-Adverse Weather}}}} &\multicolumn{5}{c}{\scalebox{0.8}{S-DGOD~\cite{wu2022single-dgod} Test Domains (mAP@0.5 in \%)}} \\ \cmidrule(r){4-8}
Methods &$\text{Params.}^{*}$ &DS &$\rightarrow$\emph{NC} &$\rightarrow$\emph{DR} &$\rightarrow$\emph{NR} &$\rightarrow$\emph{DF} &Avg. \\\midrule
\multicolumn{8}{c}{\emph{Single-source DGOD Trained on \underline{Daytime-Sunny (DS)}}} \\
\addlinespace[1pt]
\multicolumn{8}{c}{Backbone : ResNet101~\cite{resnet} / Head : Faster R-CNN~\cite{ren2015fasterrcnn}} \\
\addlinespace[1pt]
S-DGOD~\cite{wu2022single-dgod} &\textcolor{darkgray}{42.3M} &56.1 &36.6 &28.2 &16.6 &33.5 &28.7 \\
CLIP-Gap~\cite{vidit2023clipthegap} &\textcolor{darkgray}{42.3M} &51.3 &36.9 &32.3 &18.7 &38.5 &31.6 \\
OA-DG~\cite{oamix} &\textcolor{darkgray}{42.3M} &55.8 &38.0 &33.9 &16.8 &38.3 &31.8 \\
PDOC~\cite{pdod} &\textcolor{darkgray}{42.3M} &53.6 &38.5 &33.7 &19.2 &39.1 &32.6\\
UFR~\cite{liu2024unbiased} &\textcolor{darkgray}{42.3M} &58.6 &40.8 &33.2 &19.2 &\textbf{39.6} &33.2\\
DivAlign~\cite{divalign} &\textcolor{darkgray}{42.3M} &52.8 &\textbf{42.5} &\textbf{38.1} &24.1 &37.2 &35.5 \\
\cellcolor[HTML]{efefef}SoMA (Ours) &\cellcolor[HTML]{efefef}3.1M & \cellcolor[HTML]{efefef}49.3 &\cellcolor[HTML]{efefef}41.9 &\cellcolor[HTML]{efefef}37.9 &\cellcolor[HTML]{efefef}\textbf{24.5} &\cellcolor[HTML]{efefef}38.2 &\cellcolor[HTML]{efefef}\textbf{35.6} \\ \midrule
\multicolumn{8}{c}{Backbone : DINOv2-L~\cite{oquab2023dinov2} / Head : Co-DETR~\cite{codetr}} \\
\addlinespace[1pt]
Freeze &0.0M &65.0 &54.2 &55.0 &42.8 &46.9 &49.7 \\
FFT &307.3M &68.2 &57.1 &56.6 &43.1 &47.2 &51.0 \\
DoRA~\cite{dora} &5.8M &69.0 &58.7 &58.0 &45.0 &48.9 &52.7 \\
\scalebox{0.9}{AdaptFormer}~\cite{chen2022adaptformer} &6.3M &68.9 &58.8 &58.3 &44.4 &49.8 &52.8 \\
LoRA~\cite{hu2022lora} &5.5M &69.6 &\textbf{59.6} &58.1 &46.1 &49.5 &53.3 \\
\cellcolor[HTML]{efefef}SoMA (Ours) &\cellcolor[HTML]{efefef}4.9M & \cellcolor[HTML]{efefef}69.4 &\cellcolor[HTML]{efefef}59.3 &\cellcolor[HTML]{efefef}\textbf{59.3} &\cellcolor[HTML]{efefef}\textbf{47.6} &\cellcolor[HTML]{efefef}\textbf{51.0} &\cellcolor[HTML]{efefef}\textbf{54.3} \\
\bottomrule
\end{tabular}
        \vspace{-3mm}
        \caption{
        \textbf{Domain generalized object detection.}
        }\label{tab: dgod}
        \vspace{-6.5mm}
    \end{minipage}
\end{table}

\noindent \textbf{Real-world data efficiency.}
Following \cite{rein}, we evaluate models under constrained conditions using a limited set of real-world images.
Specifically, models pre-trained on synthetic datasets \emph{GTAV} + \emph{SYNTHIA} + \emph{UrbanSyn} (marked with $\ddagger$ in Tab.~\ref{tab: dgss_s2r}) are fine-tuned on 1/16 of the \emph{Cityscapes} training set.
As shown in Tab.~\ref{tab: dgss_data_efficiency}, SoMA achieves performance on par with Rein in the source domain, while offering greater robustness on unseen domains.
This underscores SoMA's data-efficient adaptability to real-world scenes, while effectively avoiding overfitting to the source domain.

\begin{figure}[t]
    \vspace{-3.4mm}
    \centering
    \includegraphics[width=0.85\linewidth, keepaspectratio]{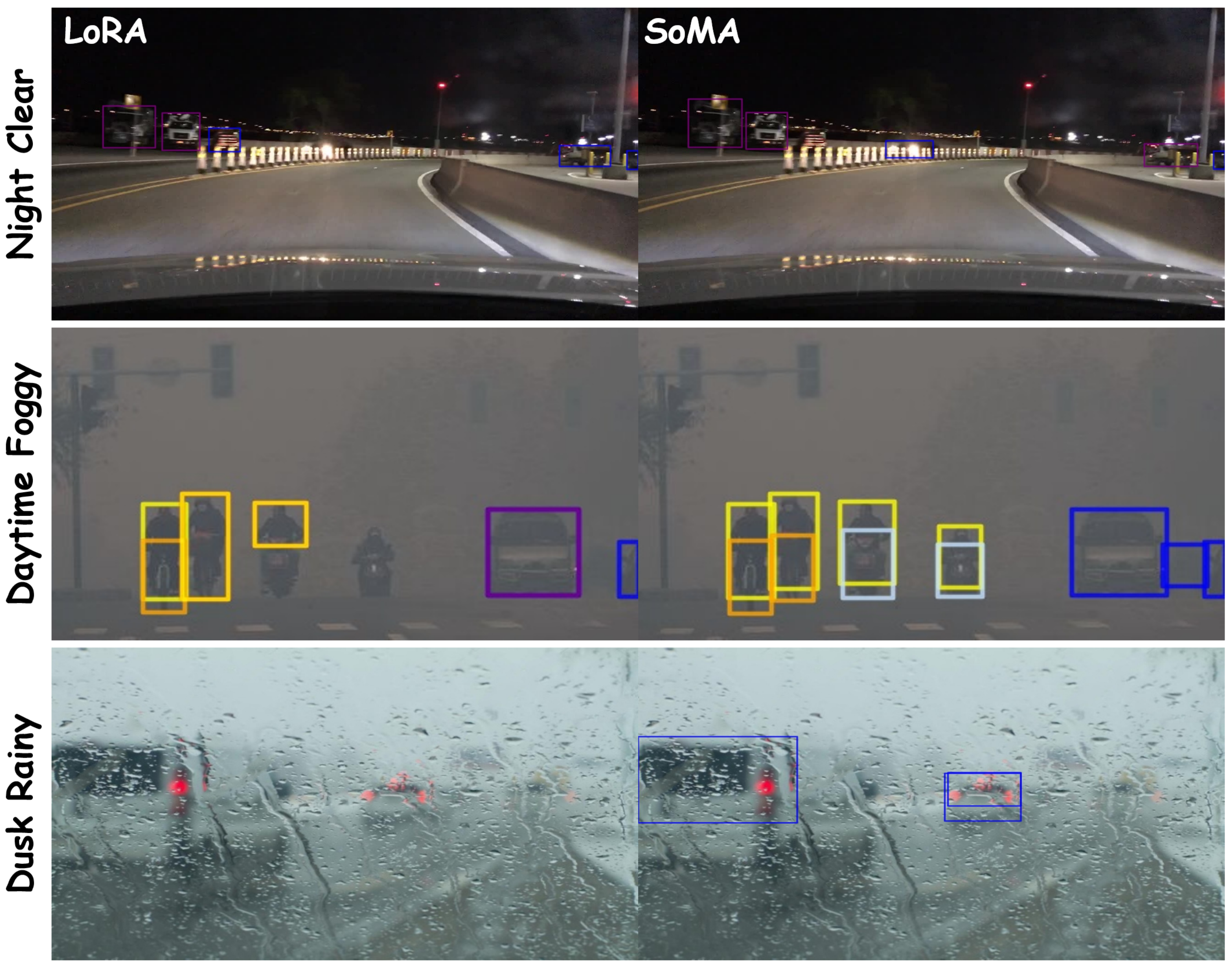}
    \vspace{-3mm}
    \caption{\textbf{DGOD qualitative results.} 
    }
    \label{fig:dgod_qualitative}
    \vspace{-1.5mm}
\end{figure}

\noindent \textbf{Clear-to-adverse weather DGOD.}
In Tab.~\ref{tab: dgod}, we compare SoMA with existing DGOD methods for classic backbone-head setting and with PEFT methods for recent foundation architectures.
As expected, SoMA demonstrates only minor improvements in the classic setting; however, in the VFM setting—where there is considerably more knowledge to retain—it shows substantial performance gains.
As illustrated by the visualization results in Fig.~\ref{fig:dgod_qualitative}, SoMA makes robust predictions even under diverse challenging scenes.

\begin{table}[t]
    \begin{minipage}{0.48\textwidth}
    \centering
        \tablestyle{2.2pt}{0.95}
        \begin{tabular}{l|ccc}  
\toprule
Methods &$\text{Params.}^{*}$ &DGSS \scalebox{0.8}{Avg.} &DGOD \scalebox{0.8}{Avg.} \\\midrule
Full fine-tuning (baseline) &304.2M &64.4 &51.0 \\
$\textbf{\raisebox{0.2em}{\(\llcorner\)}}$ + Freezing early blocks &201.6M &65.0 {\tiny \textcolor{red}{($\uparrow$ 0.6)}} &51.4 {\tiny \textcolor{red}{($\uparrow$ 0.4)}} \\
\textcolor{darkgray}{\hspace{1.2mm} $\textbf{\raisebox{0.2em}{\(\llcorner\)}}$ + Tuning principal components} \hspace{1mm} &\textcolor{darkgray}{4.9M} &\textcolor{darkgray}{66.1 {\tiny \textcolor{red}{($\uparrow$ 1.1)}}} &\textcolor{darkgray}{53.0 {\tiny \textcolor{red}{($\uparrow$ 1.6)}}} \\
\hspace{1.2mm} $\textbf{\raisebox{0.2em}{\(\llcorner\)}}$ + Tuning minor components &4.9M &67.7 {\tiny \textcolor{red}{($\uparrow$ 2.7)}} &53.8 {\tiny \textcolor{red}{($\uparrow$ 2.4)}}\\
\cellcolor[HTML]{efefef}\hspace{2.4mm} $\textbf{\raisebox{0.2em}{\(\llcorner\)}}$ + Annealing weight decay &\cellcolor[HTML]{efefef}4.9M &\cellcolor[HTML]{efefef}68.3 {\tiny \textcolor{red}{($\uparrow$ 0.6)}} &\cellcolor[HTML]{efefef}54.3 {\tiny \textcolor{red}{($\uparrow$ 0.5)}}\\
\bottomrule
\end{tabular}
        \vspace{-3mm}
        \caption{
        \textbf{Effect of our changes} evaluated on DG benchmarks. See the full ablation study in the \emph{Supplemental}.
        }\label{tab: component_ablation}
        \vspace{1.3mm}
    \end{minipage} \\
    \begin{minipage}{0.48\textwidth}
    \centering
        \tablestyle{1.6pt}{0.95}
        \begin{tabular}{cc|ccccc} 
\toprule
\scalebox{0.8}{\{$W_q$,$W_k$,$W_v$,$W_o$\}} &\scalebox{0.8}{\{$W_{up}$,$W_{down}$\}} &\scalebox{0.85}{$\text{Params.}^{*}$} &$\rightarrow$\emph{Citys.} &$\rightarrow$\emph{BDD} &$\rightarrow$\emph{Map.} &Avg. \\\midrule
\raisebox{-0.2em}{\textcolor{gray}{\CheckmarkBold}} &\raisebox{-0.2em}{\CheckmarkBold} &2.7M &68.76 &\textbf{61.50} &70.00 &66.75\\
\raisebox{-0.2em}{\CheckmarkBold} &\raisebox{-0.2em}{\textcolor{gray}{\CheckmarkBold}} &2.2M &70.82 &61.36 &69.86 &67.35\\
\cellcolor[HTML]{efefef}\raisebox{-0.2em}{\CheckmarkBold} &\cellcolor[HTML]{efefef}\raisebox{-0.2em}{\CheckmarkBold} &\cellcolor[HTML]{efefef}4.9M &\cellcolor[HTML]{efefef}\textbf{71.82} &\cellcolor[HTML]{efefef}61.31 &\cellcolor[HTML]{efefef}\textbf{71.67} &\cellcolor[HTML]{efefef}\textbf{68.27}\\
\bottomrule
\end{tabular}
        \vspace{-3mm}
        \caption{
        DGSS performance after applying SoMA to \textbf{different types of modules (Self-attention / MLP)} in DINOv2-Large.
        }\label{tab: module_ablation}
        \vspace{1.3mm}
    \end{minipage} \\
    \begin{minipage}{0.48\textwidth}
    \centering
        \tablestyle{4.8pt}{0.95}
        \begin{tabular}{l|ccccc} 
\toprule
Rank $r$ &4 &8 &\cellcolor[HTML]{efefef}16 &32 &64 \\\midrule
DGSS avg. (mIoU in \%) \hspace{1.5mm} &66.91 &67.71 &\cellcolor[HTML]{efefef}\textbf{68.27} &67.76 &67.59 \\
$\text{Params.}^{*}$ &1.3M &2.5M &\cellcolor[HTML]{efefef}4.9M &9.6M &19.0M \\
\bottomrule
\end{tabular}
        \vspace{-3mm}
        \caption{
        DGSS performance with \textbf{different rank $r$.}
        }\label{tab: rank_ablation}
        \vspace{-5mm}
    \end{minipage} 
\end{table}


\vspace{-3mm}
\subsection{Ablation Study}
\vspace{-0.5mm}

\textbf{Component Analysis.} A thorough examination of each component of SoMA in the DGOD and DGSS basic setting, as summarized in Tab.~\ref{tab: component_ablation}, demonstrates that all components of SoMA incrementally improve DG performance \emph{without adding computational costs}.
Notably, the marked performance gap with tuning principal singular components~\cite{meng2024pissa} suggests that SoMA better preserves the integrity of pre-trained representations while learning task-specific features.

\noindent \textbf{Tuning granularity.} Interestingly, as shown in Tables~\ref{tab: module_ablation} and \ref{tab: rank_ablation}, our method consistently achieves competitive results compared to SOTA baselines across all tuning granularities, indicating that SoMA can be flexibly configured according to the training budget, task difficulty, or domain gap.
The best results are achieved with the default setting.

\begin{figure}[t]
    \vspace{-3.4mm}
    \centering
    \includegraphics[width=0.7\linewidth, keepaspectratio]{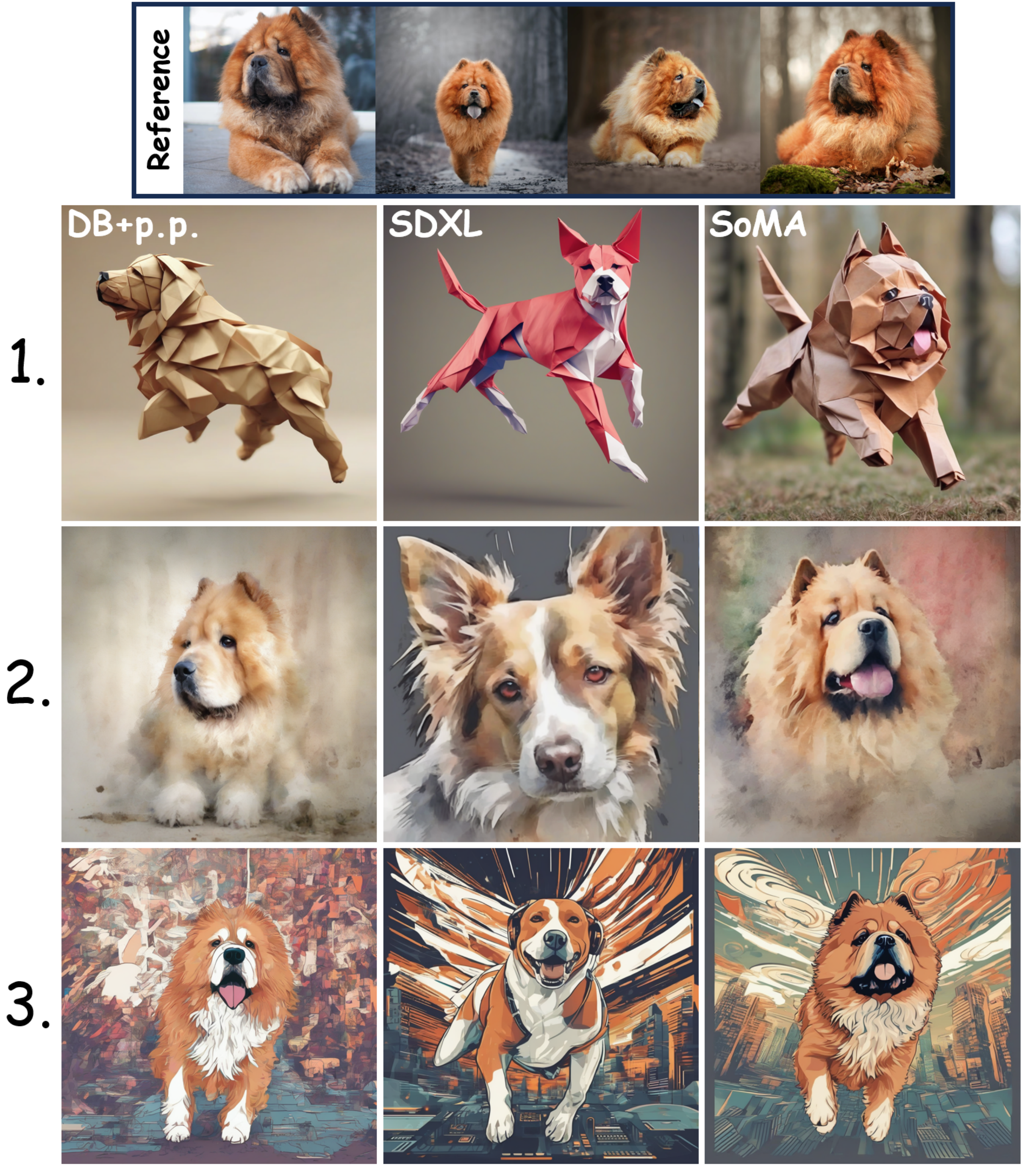}
    \vspace{-3mm}
    \caption{\textbf{Subject Personalization.} 1. A \emph{dog} gracefully leaping in origami style, 2. A \emph{dog} in watercolor painting style, and 3. A \emph{dog} soaring through a digital landscape in vector illustration style.
    }
    \label{fig:sdg}
    \vspace{-5mm}
\end{figure}

\vspace{-1mm}
\subsection{Subject Personalization}
\vspace{-1mm}
Domain generalized recognition requires consistent processing of inputs from diverse domains, whereas domain generalized generation involves generating outputs across a range of domains.
Although large-scale Text-to-Image (T2I) models have convincingly demonstrated this ability, it can be compromised in subject personalization tasks involving fine-tuning.
DreamBooth (DB)~\cite{ruiz2023dreambooth} introduces subject personalization, which fine-tunes a pre-trained T2I model on a few reference images to enable it to generate new visual concepts.
DB updates all parameters within specific blocks, potentially impairing the pre-trained capability to generate images across various domains.
Therefore, we explore whether SoMA can be seamlessly integrated into the DB framework to preserve pre-trained knowledge.
For this task, we utilize Stable Diffusion XL (SDXL)~\cite{podell2024sdxl} for the T2I diffusion model.
Fig.~\ref{fig:sdg} presents a visual comparison between DB and SoMA, using identical hyperparameter settings and sample seeds.
As seen, SoMA, similar to the original SDXL model, can generate high-quality images of diverse visual domains (e.g., style, texture, pose) and exhibits superior identity preservation compared to DB with prior preservation (p.p.) loss~\cite{ruiz2023dreambooth}.
More results with varied prompts and subjects are provided in the \emph{Supplemental}.

\vspace{-1mm}
\section{Conclusion}
\vspace{-1mm}
This paper presents a multifaceted exploration of the distribution of generalizable components in VFMs—examining them at the weight level, block level, and within training dynamics—and introduces the SoMA framework as an effective means to preserve generalizable components.
Contrary to recent approaches that passively refine frozen VFM features, SoMA is integrated into all linear layers to effectively facilitate task adaptation, while minimizing interference with generalizable components by initializing the low-rank adapter with minor singular components.
Through extensive experiments, we observe that SoMA significantly outperforms the state-of-the-art methods in DGSS, DGOD, and even subject-driven image generation tasks. 

\vspace{1mm}
\noindent\textbf{Acknowledgement:}
This work was supported by the National Research Foundation (NRF) grant funded by the Korea government (MSIT) [RS-2025-00562400] and [RS-2022-NR068754].

\appendix
\maketitlesupplementary

\noindent For a comprehensive understanding of our proposed SoMA framework, we have provided this supplementary material.
The following table of contents gives a concise overview and directs readers to specific sections of interest.

\hypersetup{linkbordercolor=black,linkcolor=black}

\renewcommand{\thesection}{\Alph{section}}
\setcounter{section}{0}

\setlength{\cftbeforesecskip}{0.5em}
\cftsetindents{section}{0em}{1.8em}
\cftsetindents{subsection}{1em}{2.5em}

\etoctoccontentsline{part}{Appendix}
\localtableofcontents
\hypersetup{linkbordercolor=blue,linkcolor=blue}

\section{Implementation Details}
We utilize the MMSegmentation~\cite{mmseg2020} and MMDetection~\cite{mmdetection} codebase for Domain Generalized Semantic Segmentation (DGSS) and Domain Generalized Object Detection (DGOD) implementations, respectively, and leverage the training scripts developed by HuggingFace~\cite{von-platen-etal-2022-diffusers} for subject personalization experiments.

\subsection{DGSS Settings}
The experimental settings for all studies conducted in the main paper are outlined in Tab.~\ref{tab: supp_training_setting}.
Unless otherwise specified, Mask2Former~\cite{cheng2022mask2former} is utilized as the default decode head, and following Rein~\cite{rein}, we adopt only the basic data augmentation used in Mask2Former.
Additionally, EMA is selectively employed to ensure stable training.
To avoid potential overfitting due to the large model (dimension) size when employing DINOv2-giant as the backbone, we opt to lower the SoMA rank from 16 to 8.

\subsection{DGOD Settings}
The DGOD settings are detailed in the rightmost two columns of Tab.~\ref{tab: supp_training_setting}.
When applying SoMA to convolution-based backbones such as ResNet~\cite{resnet}, we linearize both the patch-level convolution and its weights.
Specifically, a single convolution operation can be represented as a linear layer, $y = Wx$, where $x \in \mathbb{R}^{(n\times h\times w) \times 1}$, $y \in \mathbb{R}^{m}$, and $W \in \mathbb{R}^{m \times (n\times h \times w)}$.
We then apply SoMA as described in Eq.~\ref{equ:lora}.
For ResNet backbones, which possess a much narrower pre-trained knowledge compared to VFMs, we use extensive image corruption techniques to simulate domain shifts, following DivAlign~\cite{divalign}.
In contrast, when using DINOv2~\cite{oquab2023dinov2} as the backbone, we simply utilize basic data augmentation used in Co-DETR~\cite{codetr}.

\subsection{Subject Personalization Settings}
We conduct experiments on the DreamBooth dataset~\cite{ruiz2023dreambooth}, which consists of 30 subjects with 4–6 images per subject.
In all experiments the SoMA weights are trained using Adam optimizer for 500 iterations with a learning rate of $5e-5$.
We set the adapter rank to $r=32$ and only use a center crop for data augmentation.
Inspired by recent findings~\cite{frenkel2025b-lora} that the first 10 attention layers of \texttt{up\_blocks.0} in SDXL~\cite{podell2024sdxl} are pivotal for preserving image content, we fine-tune only these layers.
Furthermore, to fully leverage pre-trained image-text joint representations, we freeze the cross-attention modules and apply SoMA solely to the self-attention modules.

\begin{table*}[h]
    \begin{minipage}{0.95\textwidth}
    \centering
        \tablestyle{2.4pt}{1.05}
        \begin{tabular}{lcccccccc}\toprule
Hyperparameters &\multicolumn{6}{c}{DGSS} & \multicolumn{2}{c}{DGOD} \\
\addlinespace[2pt]
Setting &\emph{G}$\rightarrow$\{\emph{C, B, M}\} &\emph{G+S}$\rightarrow$\{\emph{C, B, M}\} &\multicolumn{2}{c}{\emph{G+S+U}$\rightarrow$\{\emph{C, B, M}\}} &\emph{C}$\rightarrow$\{\emph{B, M}\}\emph{/ACDC} &\emph{$\frac{1}{16}$C}$\rightarrow$\{\emph{C, B, M}\} &\multicolumn{2}{c}{\emph{DS}$\rightarrow$\{\emph{NC, DR, NR, DF}\}} \\
\addlinespace[1pt]
\cmidrule{4-5}  \cmidrule{8-9}
\addlinespace[2pt]
Backbone &DINOv2-L/EVA02-L &DINOv2-L &DINOv2-L &DINOv2-G &DINOv2-L &DINOv2-L &DINOv2-L &RN101  \\\midrule
rank $r$ &16 &16 &16 &8 &16 &16 &16 &24 \\
NFEB &8 &8 &8 &12 &8 &8 &8 &21 \\
optimizer &\multicolumn{8}{c}{AdamW} \\
lr scheduler & Linear &Linear &Linear &Linear &Linear &Linear &MultiStep &MultiStep \\
AWD scheduler &\multicolumn{8}{c}{Cosine} \\
learning rate &1e-4 &1e-4 &1e-4 &1e-4 &1e-4 &1e-5 &2e-4 &2e-4 \\
backbone lr mult. &\multicolumn{8}{c}{0.5} \\
weight decay &5e-2/3e-2 &5e-2 &5e-2 &5e-2 &5e-2 &5e-2 &5e-2 &1e-3 \\
batch size &4 &4 &4 &4 &8 &8 &8 &4 \\
warmup iters &0 &0 &0 &0 &1.5k/10k &0 &1.5k &1.5k \\
iters &40k &40k &40k &40k &40k &4k &40k &40k \\
EMA &\checkmark &\textcolor{gray}{\checkmark} &\textcolor{gray}{\checkmark} &\checkmark &\checkmark &\textcolor{gray}{\checkmark} &\textcolor{gray}{\checkmark} &\textcolor{gray}{\checkmark} \\
\bottomrule
\end{tabular}

        \vspace{-1mm}
        \caption{
        \textbf{DGSS/DGOD hyperparameter configurations.} “NFEB” denotes the number of frozen early blocks.
        }\label{tab: supp_training_setting}
        \vspace{1.5mm} 
    \end{minipage} 
\end{table*}

\begin{table*}[h]
    \begin{minipage}{0.95\textwidth}
        \centering
        \tablestyle{1.1pt}{1.1}
        \begin{tabular}{l|r|ccccccccccccccccccc|c}  
\toprule
Methods &$\text{Params.}$ &road &side. &build. &wall &fence &pole &light &sign &vege. &terr. &sky &pers. &rider &car &truck &bus &train &motor. &bicy. &mIoU \\\midrule
Full fine-tuning (baseline) &304.2M &92.1 &64.5 &87.8 &49.0 &\textbf{56.4} &58.8 &66.1 &57.3 &\textbf{82.9} &\textbf{53.9} &\textbf{95.0} &79.3 &\textbf{63.2} &\underline{91.8} &65.3 &75.9 &50.9 &64.7 &54.1 &68.9 \\
$\textbf{\raisebox{0.2em}{\(\llcorner\)}}$ + Freezing early blocks &201.6M &91.6 &65.0 &87.6 &46.9 &\underline{55.0} &57.0 &66.6 &52.9 &81.5 &52.5 &94.5 &77.9 &56.4 &91.6 &64.5 &75.8 &60.1 &68.8 &57.3 &68.6 \\
\textcolor{darkgray}{\hspace{1.2mm} $\textbf{\raisebox{0.2em}{\(\llcorner\)}}$ + Tuning principal components} &\textcolor{darkgray}{4.9M} &\textcolor{darkgray}{93.1} &\textcolor{darkgray}{69.4} &\textcolor{darkgray}{\underline{88.3}} &\textcolor{darkgray}{48.4} &\textcolor{darkgray}{53.9} &\textcolor{darkgray}{\underline{59.0}} &\textcolor{darkgray}{67.4} &\textcolor{darkgray}{57.5} &\textcolor{darkgray}{\underline{81.9}} &\textcolor{darkgray}{\underline{53.1}} &\textcolor{darkgray}{94.8} &\textcolor{darkgray}{\underline{79.9}} &\textcolor{darkgray}{\underline{61.4}} &\textcolor{darkgray}{90.0} &\textcolor{darkgray}{65.6} &\textcolor{darkgray}{80.8} &\textcolor{darkgray}{44.7} &\textcolor{darkgray}{70.7} &\textcolor{darkgray}{55.3} &\textcolor{darkgray}{69.2} \\
\hspace{1.2mm} $\textbf{\raisebox{0.2em}{\(\llcorner\)}}$ + Tuning minor components &4.9M &\underline{93.4} &\underline{70.6} &88.2 &\textbf{52.4} &55.0 &\textbf{59.1} &\underline{68.3} &\underline{60.4} &81.8 &52.3 &94.9 &\textbf{79.9} &61.0 &91.2 &\underline{69.7} &\underline{84.5} &\underline{60.2} &\underline{70.9} &\underline{59.3} &\underline{71.2}\\
\cellcolor[HTML]{efefef}\hspace{2.4mm} $\textbf{\raisebox{0.2em}{\(\llcorner\)}}$ + Annealing weight decay &\cellcolor[HTML]{efefef}4.9M &\cellcolor[HTML]{efefef}\textbf{93.6} &\cellcolor[HTML]{efefef}\textbf{71.4} &\cellcolor[HTML]{efefef}\textbf{88.3} &\cellcolor[HTML]{efefef}\underline{52.3} &\cellcolor[HTML]{efefef}54.9 &\cellcolor[HTML]{efefef}\textbf{59.1} &\cellcolor[HTML]{efefef}\textbf{69.4} &\cellcolor[HTML]{efefef}\textbf{62.7} &\cellcolor[HTML]{efefef}\underline{81.9} &\cellcolor[HTML]{efefef}52.8 &\cellcolor[HTML]{efefef}\underline{94.9} &\cellcolor[HTML]{efefef}79.4 &\cellcolor[HTML]{efefef}57.7 &\cellcolor[HTML]{efefef}\textbf{91.8} &\cellcolor[HTML]{efefef}\textbf{72.9} &\cellcolor[HTML]{efefef}\textbf{85.3} &\cellcolor[HTML]{efefef}\textbf{61.9} &\cellcolor[HTML]{efefef}\textbf{71.6} &\cellcolor[HTML]{efefef}\textbf{60.1} &\cellcolor[HTML]{efefef}\textbf{71.7}\\
\bottomrule
\end{tabular}
        \vspace{-1mm}
        \caption{
        \textbf{Effect of the proposed components under \emph{GTAV} $\rightarrow$ \emph{Mapillary} DGSS setting}. We highlight the \textbf{best} and \underline{second-best} for each column.
        }\label{tab: supp_dgss_component}
        \vspace{3mm} 
    \end{minipage} \\
    \begin{minipage}{0.95\textwidth}
        \centering
        \tablestyle{2.5pt}{1.1}
        \begin{tabular}{l|r|cccccccc|cccccccc}  
\toprule
\multicolumn{2}{c}{} &\multicolumn{8}{c}{\emph{Daytime-Sunny}$\rightarrow$\emph{Dusk-Rainy}} &\multicolumn{8}{c}{\emph{Daytime-Sunny}$\rightarrow$\emph{Daytime-Foggy}} \\
\cmidrule(lr){3-10} \cmidrule(lr){11-18}
Methods &$\text{Params.}$ &bus &bike &car &motor &person &rider &truck &mAP &bus &bike &car &motor &person &rider &truck &mAP \\\midrule
Full fine-tuning (baseline) &307.3M &61.2 &44.9 &80.7 &45.2 &56.0 &41.0 &67.2 &56.6 &46.4 &37.3 &65.1 &42.9 &49.2 &48.9 &40.9 &47.2 \\
$\textbf{\raisebox{0.2em}{\(\llcorner\)}}$ + Freezing early blocks &201.6M &63.0 &46.5 &\textbf{81.2} &46.0 &58.2 &39.5 &68.0 &57.5 &47.5 &39.5 &67.2 &45.6 &50.3 &50.0 &40.6 &48.7 \\
\textcolor{darkgray}{\hspace{1.2mm} $\textbf{\raisebox{0.2em}{\(\llcorner\)}}$ + Tuning principal components} \hspace{2mm} &\textcolor{darkgray}{4.9M} &\textcolor{darkgray}{64.3} &\textcolor{darkgray}{49.2} &\textcolor{darkgray}{80.5} &\textcolor{darkgray}{46.3} &\textcolor{darkgray}{57.9} &\textcolor{darkgray}{\underline{41.6}} &\textcolor{darkgray}{68.2} &\textcolor{darkgray}{58.3} &\textcolor{darkgray}{48.5} &\textcolor{darkgray}{\underline{40.1}} &\textcolor{darkgray}{67.5} &\textcolor{darkgray}{47.5} &\textcolor{darkgray}{50.3} &\textcolor{darkgray}{51.0} &\textcolor{darkgray}{45.3} &\textcolor{darkgray}{50.0} \\
\hspace{1.2mm} $\textbf{\raisebox{0.2em}{\(\llcorner\)}}$ + Tuning minor components &4.9M &\underline{65.6} &\textbf{50.6} &80.8 &\underline{46.9} &\underline{58.4} &\textbf{42.7} &\underline{69.4} &\underline{59.2} &\textbf{50.6} &39.7 &\underline{67.7} &\underline{48.3} &\underline{50.9} &\underline{51.6} &\textbf{46.1} &\underline{50.7} \\
\cellcolor[HTML]{efefef}\hspace{2.4mm} $\textbf{\raisebox{0.2em}{\(\llcorner\)}}$ + Annealing weight decay &\cellcolor[HTML]{efefef}4.9M &\cellcolor[HTML]{efefef}\textbf{65.7} &\cellcolor[HTML]{efefef}\underline{50.5} &\cellcolor[HTML]{efefef}\underline{81.1} &\cellcolor[HTML]{efefef}\textbf{48.1} &\cellcolor[HTML]{efefef}\textbf{59.0} &\cellcolor[HTML]{efefef}41.0 &\cellcolor[HTML]{efefef}\textbf{69.5} &\cellcolor[HTML]{efefef}\textbf{59.3} &\cellcolor[HTML]{efefef}\underline{50.3} &\cellcolor[HTML]{efefef}\textbf{40.9} &\cellcolor[HTML]{efefef}\textbf{67.9} &\cellcolor[HTML]{efefef}\textbf{48.6} &\cellcolor[HTML]{efefef}\textbf{51.5} &\cellcolor[HTML]{efefef}\textbf{52.3} &\cellcolor[HTML]{efefef}\underline{45.7} &\cellcolor[HTML]{efefef}\textbf{51.0}\\
\bottomrule
\end{tabular}
        \vspace{-1mm}
        \caption{
        \textbf{Effect of the proposed components under \emph{Daytime-Sunny} $\rightarrow$ \emph{Dusk-Rainy} and \emph{Daytime-Sunny} $\rightarrow$ \emph{Daytime-Foggy} DGOD settings}. We highlight the \textbf{best} and \underline{second-best} for each column.
        }\label{tab: supp_dgod_component}
        \vspace{-2mm} 
    \end{minipage}
\end{table*}

\vspace{-1.5mm}
\section{Detailed Ablations}
\vspace{-0.5mm}
\subsection{Component Analysis} \label{supp_sec: ablation}
In this subsection, we conduct detailed ablation studies under multiple settings: \emph{GTAV} $\rightarrow$ \emph{Mapillary} DGSS and \emph{Daytime-Sunny} $\rightarrow$ \{\emph{Dusk-Rainy, Daytime-Foggy}\} DGOD scenarios.
In Tables \ref{tab: supp_dgss_component} and \ref{tab: supp_dgod_component}, we systematically evaluate the effectiveness of each component within the SoMA framework based on class-wise IoU/AP (\%).
All proposed components enhance overall generalization performance without adding any additional training or inference costs.

As illustrated in Tab.~\ref{tab: supp_dgss_component}, \emph{freezing early blocks} not only substantially reduces the number of trainable parameters but also significantly improves performance for classes that are infrequently observed in the source dataset (e.g., bicycle, motorcycle, train).
Additionally, \emph{tuning minor singular components} maximizes the retention of VFM's world knowledge during task adaptation, leading to superior generalization performance over tuning principal components (PiSSA~\cite{meng2024pissa}) for most classes.
For an in-depth comparison of our methods with PiSSA, please refer to Sec.~\ref{sec: add. comp.}.
Lastly, \emph{annealing weight decay} proves especially beneficial for classes requiring fine-detail discrimination (e.g., road \emph{vs.}\hspace{0.8mm}sidewalk, traffic light \emph{vs.}\hspace{0.8mm}traffic sign, car \emph{vs.}\hspace{0.8mm}truck \emph{vs.}\hspace{0.8mm}bus \emph{vs.}\hspace{0.8mm}train, motorcycle \emph{vs.}\hspace{0.8mm}bicycle).
Likewise, all components clearly improve recognition for the majority of classes under adverse weather detection settings (see Tab.~\ref{tab: supp_dgod_component}).

\subsection{Freezing Scheme}
While freezing the initial blocks of VFM is effective in preserving its generalization ability during task adaptation, freezing too many blocks can lead to a reduction in discriminability.
To better understand this trade-off, we explore the effects of varying the number of frozen blocks.
Tab.~\ref{tab: supp_nfeb_ablation} shows that freezing up to the first 8 blocks progressively enhances performance, but freezing beyond this point results in a decline.
Considering that feature maps from multiple blocks (e.g., the 8th, 12th, 16th, and 24th blocks in large-sized backbones) serve as inputs to the segmentation/detection head, using more than one frozen VFM features as head input significantly undermines task adaptability (\emph{i.e.} discriminability).
\begin{table}[t]
    \begin{minipage}{0.48\textwidth}
    \centering
        \tablestyle{4.4pt}{1.1}
        \begin{tabular}{l|ccccc} 
\toprule
\# frozen early blocks &0 &4 &\cellcolor[HTML]{efefef}8 &12 &16 \\\midrule
\emph{Citys.} perf. (mIoU in \%) \hspace{1.5mm} &70.62 &71.51 &\cellcolor[HTML]{efefef}\textbf{71.82} &70.71 &70.47 \\
$\text{Params.}^{*}$ &7.3M &6.1M &\cellcolor[HTML]{efefef}4.9M &3.7M &2.4M \\
\bottomrule
\end{tabular}
        \vspace{-1mm}
        \caption{
        Performance comparison with \textbf{varying numbers of frozen early blocks} under \emph{GTAV} $\rightarrow$ \emph{Cityscapes} DGSS setting.
        }\label{tab: supp_nfeb_ablation}
        \vspace{-6mm} 
    \end{minipage} 
\end{table}
Furthermore, since the first input feature map of the decode head is directly incorporated into the final mask prediction in Mask2Former~\cite{cheng2022mask2former}, freezing the blocks that generate this feature map allows the full utilization of the generalization capacity of the early blocks in VFMs (see Fig.~\ref{fig:early_freeze}).

\section{Additional Experiments}

\subsection{Results on Various Backbones}
Tab.~\ref{tab: supp_various_backbone} showcases the versatility of SoMA across a wide range of backbones, ranging from isotropic Vision Transformers (ViTs) to ConvNets and hierarchical ViT, as well as models trained under various approaches, such as ImageNet~\cite{deng2009imagenet} supervision and MAE~\cite{he2022masked, convnextv2} pre-training.
SoMA consistently outperforms FFT across diverse backbone architectures.
Notably, the improvements brought by SoMA become increasingly pronounced with larger model sizes and more extensive, high-quality data during pre-training, highlighting the superior ability of our method to preserve pre-trained knowledge.

\begin{table}[t]
    \begin{minipage}{0.48\textwidth}
    \centering
        \tablestyle{1.6pt}{1.1}
        \begin{tabular}{llccccc} 
\toprule
\multicolumn{3}{c}{\textbf{\emph{Backbone Ablation}}}&\multicolumn{4}{c}{Test Domains (mIoU in \%)} \\ \cmidrule(r){4-7}
Backbones & Methods &$\text{Params.}^{*}$ &$\rightarrow$\emph{Citys.} &$\rightarrow$\emph{BDD} &$\rightarrow$\emph{Map.} &Avg. \\\midrule
\multicolumn{7}{c}{\emph{Single-source DGSS Trained on \underline{GTAV}}} \\
\addlinespace[2pt] 
\multirow{2}{*}{DINOv2-L~\cite{oquab2023dinov2}} &FFT &304.2M &66.93 &57.34 &68.89 &64.39 \\
&\cellcolor[HTML]{efefef}SoMA  &\cellcolor[HTML]{efefef}4.9M &\cellcolor[HTML]{efefef}\textbf{71.82} &\cellcolor[HTML]{efefef}\textbf{61.31} &\cellcolor[HTML]{efefef}\textbf{71.67} &\cellcolor[HTML]{efefef}\textbf{68.27} \\ \midrule
\multirow{2}{*}{DINOv2-B~\cite{oquab2023dinov2}} &FFT &86.5M &60.84 &52.98 &62.12 &58.65 \\
&\cellcolor[HTML]{efefef}SoMA  &\cellcolor[HTML]{efefef}2.3M &\cellcolor[HTML]{efefef}\textbf{66.71} &\cellcolor[HTML]{efefef}\textbf{57.48} &\cellcolor[HTML]{efefef}\textbf{67.34} &\cellcolor[HTML]{efefef}\textbf{63.84} \\ \midrule
\multirow{2}{*}{DINOv2-S~\cite{oquab2023dinov2}} &FFT &22.0M &53.71 &49.03 &58.10 &53.61 \\
&\cellcolor[HTML]{efefef}SoMA  &\cellcolor[HTML]{efefef}1.0M &\cellcolor[HTML]{efefef}\textbf{57.58} &\cellcolor[HTML]{efefef}\textbf{52.95} &\cellcolor[HTML]{efefef}\textbf{62.48} &\cellcolor[HTML]{efefef}\textbf{57.67} \\ \midrule
\multirow{2}{*}{\scalebox{0.85}{ConvNeXt V2-L}~\cite{convnextv2}} &FFT &196.4M &55.93 &50.71 &60.79 &55.81 \\
&\cellcolor[HTML]{efefef}SoMA  &\cellcolor[HTML]{efefef}12.1M &\cellcolor[HTML]{efefef}\textbf{60.12} &\cellcolor[HTML]{efefef}\textbf{53.36} &\cellcolor[HTML]{efefef}\textbf{61.46} &\cellcolor[HTML]{efefef}\textbf{58.31} \\ \midrule
\multirow{2}{*}{Swin-L~\cite{liu2021swin}} &FFT &195.2M &54.40 &49.85 &60.05 &54.77 \\
&\cellcolor[HTML]{efefef}SoMA  &\cellcolor[HTML]{efefef}5.4M &\cellcolor[HTML]{efefef}\textbf{56.91} &\cellcolor[HTML]{efefef}\textbf{51.98} &\cellcolor[HTML]{efefef}\textbf{60.73} &\cellcolor[HTML]{efefef}\textbf{56.54} \\ \midrule
\multirow{2}{*}{ResNet101~\cite{resnet}} &FFT &42.3M &\textbf{41.29} &44.29 &48.79 &44.79 \\
&\cellcolor[HTML]{efefef}SoMA  &\cellcolor[HTML]{efefef}2.5M &\cellcolor[HTML]{efefef}41.23 &\cellcolor[HTML]{efefef}\textbf{45.57} &\cellcolor[HTML]{efefef}\textbf{49.71} &\cellcolor[HTML]{efefef}\textbf{45.50} \\
\bottomrule
\end{tabular}
        \vspace{-1mm}
        \caption{
        \textbf{Results across various backbones and model sizes.}
        }\label{tab: supp_various_backbone}
        \vspace{2.5mm} 
    \end{minipage} \\
    \begin{minipage}{0.48\textwidth}
    \centering
        \tablestyle{1.6pt}{1.1}
        \begin{tabular}{llccccc} 
\toprule
\multicolumn{3}{c}{\textbf{\emph{SemFPN Results}}}&\multicolumn{4}{c}{Test Domains (mIoU in \%)} \\ \cmidrule(r){4-7}
Backbones & Methods &$\text{Params.}^{*}$ &$\rightarrow$\emph{Citys.} &$\rightarrow$\emph{BDD} &$\rightarrow$\emph{Map.} &Avg. \\\midrule
\multicolumn{7}{c}{\emph{Single-source DGSS Trained on \underline{GTAV}}} \\
\addlinespace[2pt]
\multirow{2}{*}{DINOv2-L~\cite{oquab2023dinov2} \hspace{2.1mm}} &Rein~\cite{rein} &2.5M &63.60 &59.00 &63.70 &62.10 \\
&\cellcolor[HTML]{efefef}SoMA  &\cellcolor[HTML]{efefef}4.9M &\cellcolor[HTML]{efefef}\textbf{67.81} &\cellcolor[HTML]{efefef}\textbf{60.12} &\cellcolor[HTML]{efefef}\textbf{68.95} &\cellcolor[HTML]{efefef}\textbf{65.63} \\ \midrule
\multirow{2}{*}{EVA02-L~\cite{eva02}} &Rein~\cite{rein} &2.5M &61.40 &\textbf{58.50} &62.00 &60.70 \\
&\cellcolor[HTML]{efefef}SoMA  &\cellcolor[HTML]{efefef}5.1M &\cellcolor[HTML]{efefef}\textbf{64.91} &\cellcolor[HTML]{efefef}57.54 &\cellcolor[HTML]{efefef}\textbf{65.33} &\cellcolor[HTML]{efefef}\textbf{62.59} \\ 
\bottomrule
\end{tabular}
        \vspace{-1mm}
        \caption{
        DGSS evaluation results with \textbf{SemFPN head}~\cite{semfpn}.
        }\label{tab: supp_semfpn}
        \vspace{-4mm} 
    \end{minipage}
\end{table}

\subsection{Results on SemFPN Head}
While Mask2Former~\cite{cheng2022mask2former} is predominantly employed as decode head in all DGSS experiments, SoMA is compatible with any decode head.
To assess its robustness across different heads, we employ the lightweight SemFPN~\cite{semfpn} head to benchmark its performance against Rein~\cite{rein}.
Our experimental results (Tab.~\ref{tab: supp_semfpn}) indicate that SoMA integrates seamlessly with diverse backbones and heads, consistently surpassing the SOTA baseline.

\section{Model Efficiency Comparison} \label{sec: effi. comp.}
As detailed in Tab.~\ref{tab: model_efficiency}, SoMA exhibits higher throughput than adapter- and VPT-based methods like Rein~\cite{rein} and SET~\cite{set}, as it incurs no additional latency.
This advantage is especially significant in online inference settings, where the batch size is typically as small as one~\cite{hu2022lora}.
Furthermore, in scenarios involving DINOv2-giant exceeding 1B parameters, SoMA can drastically reduce training costs compared to FFT.
SoMA initialization is completed within 30 seconds for large-sized models, which is a negligible cost given the improved performance.

\section{Additional Comparison} \label{sec: add. comp.}
In Tables \ref{tab: supp_full_dgss_s2r}, \ref{tab: supp_full_dgss_r2r}, and \ref{tab: supp_dgss_acdc_val}, we present an exhaustive comparison with existing methods to illustrate the broader research landscape across multiple DGSS settings.

\begin{table}[t]
    \begin{minipage}{0.48\textwidth}
    \centering
        \tablestyle{2.6pt}{1.0}
        \begin{tabular}{lcccc}  
\toprule
\textbf{\emph{Efficiency}}&\multicolumn{2}{c}{Training (bs = 4)} &\multicolumn{2}{c}{Inference (bs = 1 / 32)} \\ \cmidrule(lr){2-3} \cmidrule(lr){4-5}
Methods &Time (hrs) &Memory &Throughput (imgs/s) &Memory \\\midrule
$\text{SET}_{\hspace{2.9mm}\text{large}}$ &9.2 &12.5G &20.0 / \hspace{2mm}-\hspace{2.3mm} &5.5G / OOM  \\
$\text{Rein}_{\hspace{2.6mm}\text{large}}$ \hspace{1.3mm} &9.3 &\textbf{12.2G} &33.6 / 64.3 &4.7G / 48.2G \\
\cellcolor[HTML]{efefef}$\text{SoMA}_{\hspace{0.5mm}\text{large}}$  &\cellcolor[HTML]{efefef}\textbf{9.0} &\cellcolor[HTML]{efefef}12.7G &\cellcolor[HTML]{efefef}\textbf{56.4 / 79.7} &\cellcolor[HTML]{efefef}\textbf{4.4G / 40.9G} \\
$\text{FFT}_{\hspace{3.2mm}\text{giant}}$ &27.9 &45.3G &21.6 / \hspace{2mm}-\hspace{2.3mm} &10.6G / OOM \\
\cellcolor[HTML]{efefef}$\text{SoMA}_{\hspace{0.5mm}\text{giant}}$  &\cellcolor[HTML]{efefef}\textbf{18.9} &\cellcolor[HTML]{efefef}\textbf{25.6G} &\cellcolor[HTML]{efefef}21.6 / \hspace{2mm}-\hspace{2.3mm} &\cellcolor[HTML]{efefef}10.6G / OOM \\
\bottomrule
\end{tabular}
        \vspace{-1mm}
        \caption{
        \textbf{DGSS model efficiency.}
        Inference statistics are measured only for the backbone on image crops of 512$\times$512, and are measured with warmup and  averaged over multiple runs. We use an NVIDIA RTX A6000.  “bs” denotes batch size.
        }\label{tab: model_efficiency}
        \vspace{1.5mm}
    \end{minipage}
    \begin{minipage}{0.48\textwidth}
    \centering
        \tablestyle{1.6pt}{1.1}
        \begin{tabular}{lcccccc} 
\toprule
\multicolumn{3}{c}{\textbf{\emph{Synthetic-to-Real Generalization}}}&\multicolumn{4}{c}{Test Domains (mIoU in \%)} \\ \cmidrule(r){4-7}
Methods &Backbone &$\text{Params.}^{*}$ &$\rightarrow$\emph{Citys.} &$\rightarrow$\emph{BDD} &$\rightarrow$\emph{Map.} &Avg. \\\midrule
\multicolumn{7}{c}{\emph{Single-source DGSS Trained on \underline{GTAV}}} \\
\addlinespace[2pt]
PEGO~\cite{pego} &DINOv2-L &2.6M &68.86 &\textbf{61.44} & 68.61 & 66.30 \\
PiSSA$_{r16}$~\cite{meng2024pissa} & DINOv2-L & 7.3M & 69.43 &60.62 & 69.44 &66.50 \\
\cellcolor[HTML]{efefef}SoMA~(Ours) &\cellcolor[HTML]{efefef}DINOv2-L & \cellcolor[HTML]{efefef}4.9M &\cellcolor[HTML]{efefef}\textbf{71.82} &\cellcolor[HTML]{efefef}61.31 &\cellcolor[HTML]{efefef}\textbf{71.67} &\cellcolor[HTML]{efefef}\textbf{68.27} \\
\bottomrule
\end{tabular}
        \vspace{-1mm}
        \caption{
        Performance Comparison of our \textbf{SoMA against PEGO and PiSSA} under the basic DGSS setting. The reported performance of PEGO indicates the best result achieved within the hyperparameter space proposed in the original paper.
        }\label{tab: pego pissa}
        \vspace{-4mm} 
    \end{minipage}
\end{table}

\noindent\textbf{Comparing SoMA with PiSSA}~\cite{meng2024pissa} \textbf{and PEGO}~\cite{pego}.
PiSSA optimizes parameter efficiency by selectively adjusting the principal singular direction, which is the most stretched direction of the weight matrix.
Also PEGO enhances domain generalization by enforcing strict orthogonality between the LoRA adapter and every direction of the pre-trained weights.
In stark contrast to PiSSA, SoMA tunes minor singular components, effectively preserving the integrity of pre-trained knowledge.
Importantly, our method accounts not only for component orthogonality but also for how pre-trained knowledge is structured within the weight matrix.
Thus, although both methods maintain similar orthogonality between tuned and frozen components, SoMA, unlike PiSSA—which directly tunes principal components—robustly adjusts the hierarchical world knowledge structure (see Fig.~2 in the main paper), achieving superior DGSS performance, as shown in Tab.~\ref{tab: pego pissa}.
These findings are consistent with the observations presented in the Ablation Sec.~\ref{supp_sec: ablation}.

Furthermore, unlike PEGO, our SoMA leverages spectral information to initialize the LoRA adapter, allowing it to naturally preserve the pre-trained knowledge structure without relying on explicit regularization loss.
As evidenced by its superior performance relative to PEGO in Tab.~\ref{tab: pego pissa}, we argue that imposing strict orthogonality constraints on all directions of the pre-trained weights may excessively restrict task adaptation, potentially compromising discriminability.
Lastly, whereas PEGO and PiSSA explore adaptation solely at the weight level, our SoMA framework extends its analysis to both the block level and training dynamics.

\begin{figure*}[h]
    \vspace{-1.5mm}
    \centering
    \includegraphics[width=\textwidth]{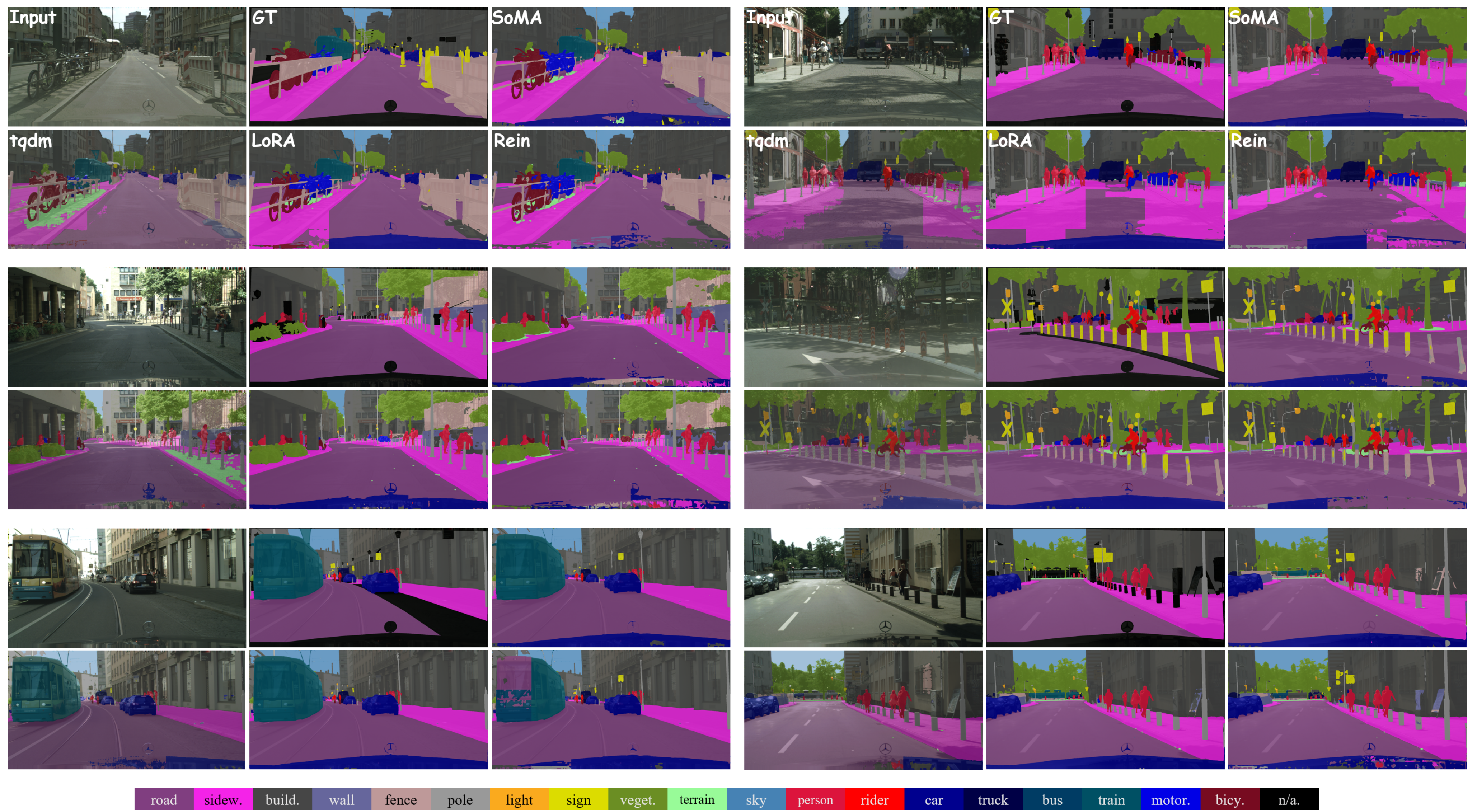}
    \vspace{-3.5mm}
    \caption{
    Segmentation results of SoMA on the Cityscapes. The model is trained on GTAV with DINOv2-L backbone.
    }
    \label{fig:supp_citys_qualitative}
    \vspace{-1mm}
\end{figure*}

\begin{figure*}[h]
    \vspace{-1.5mm}
    \centering
    \includegraphics[width=\textwidth]{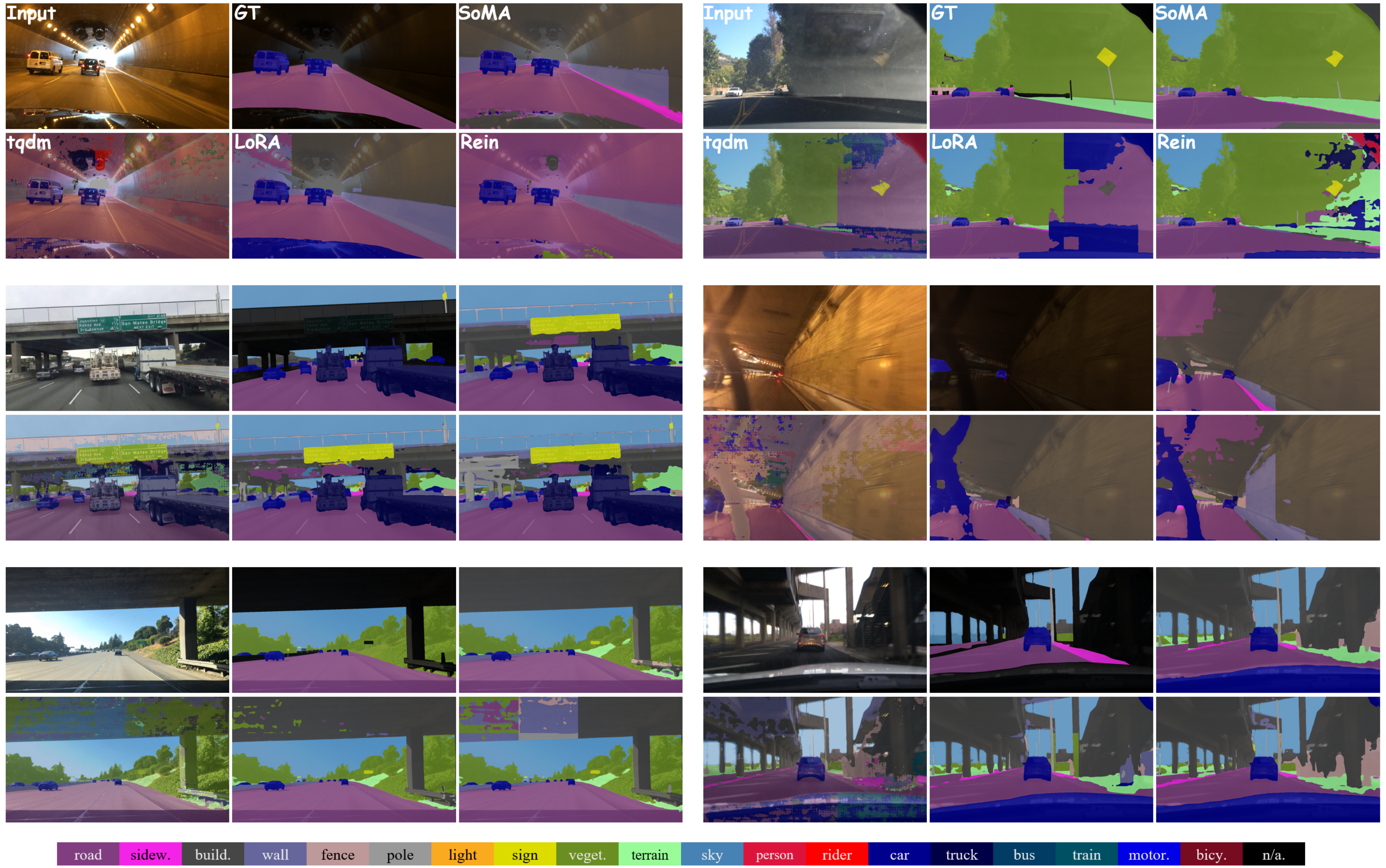}
    \vspace{-3.5mm}
    \caption{
    Segmentation results of SoMA on the BDD100k. The model is trained on GTAV with DINOv2-L backbone.
    }
    \label{fig:supp_bdd_qualitative}
    \vspace{-3mm}
\end{figure*}

\begin{figure*}[h]
    \vspace{-1.5mm}
    \centering
    \includegraphics[width=\textwidth]{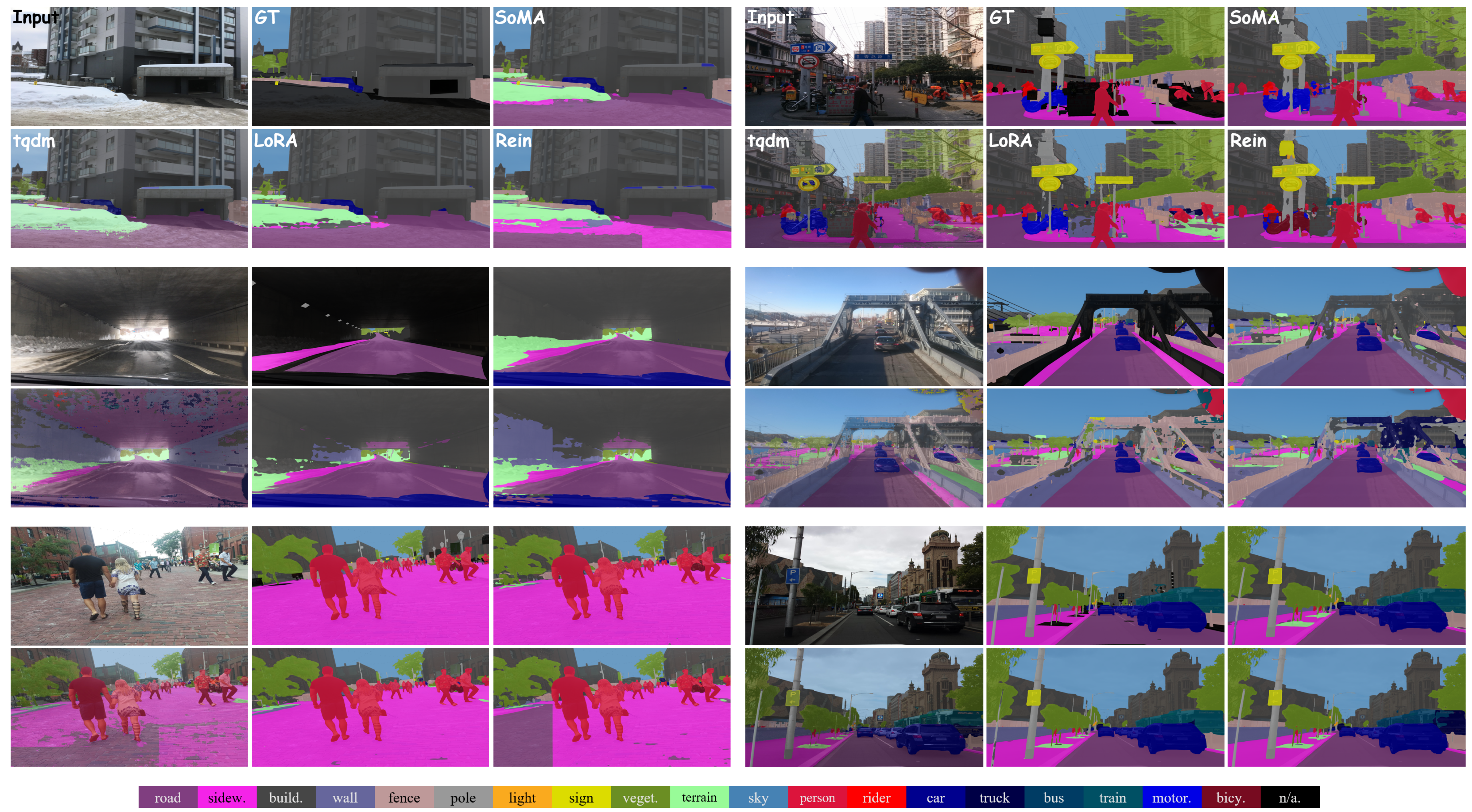}
    \vspace{-2.5mm}
    \caption{
    Segmentation results of SoMA on the Mapillary. The model is trained on GTAV with DINOv2-L backbone.
    }
    \label{fig:supp_map_qualitative}
    \vspace{2mm}
\end{figure*}

\begin{figure*}[h]
    \vspace{-1.5mm}
    \centering
    \includegraphics[width=\textwidth]{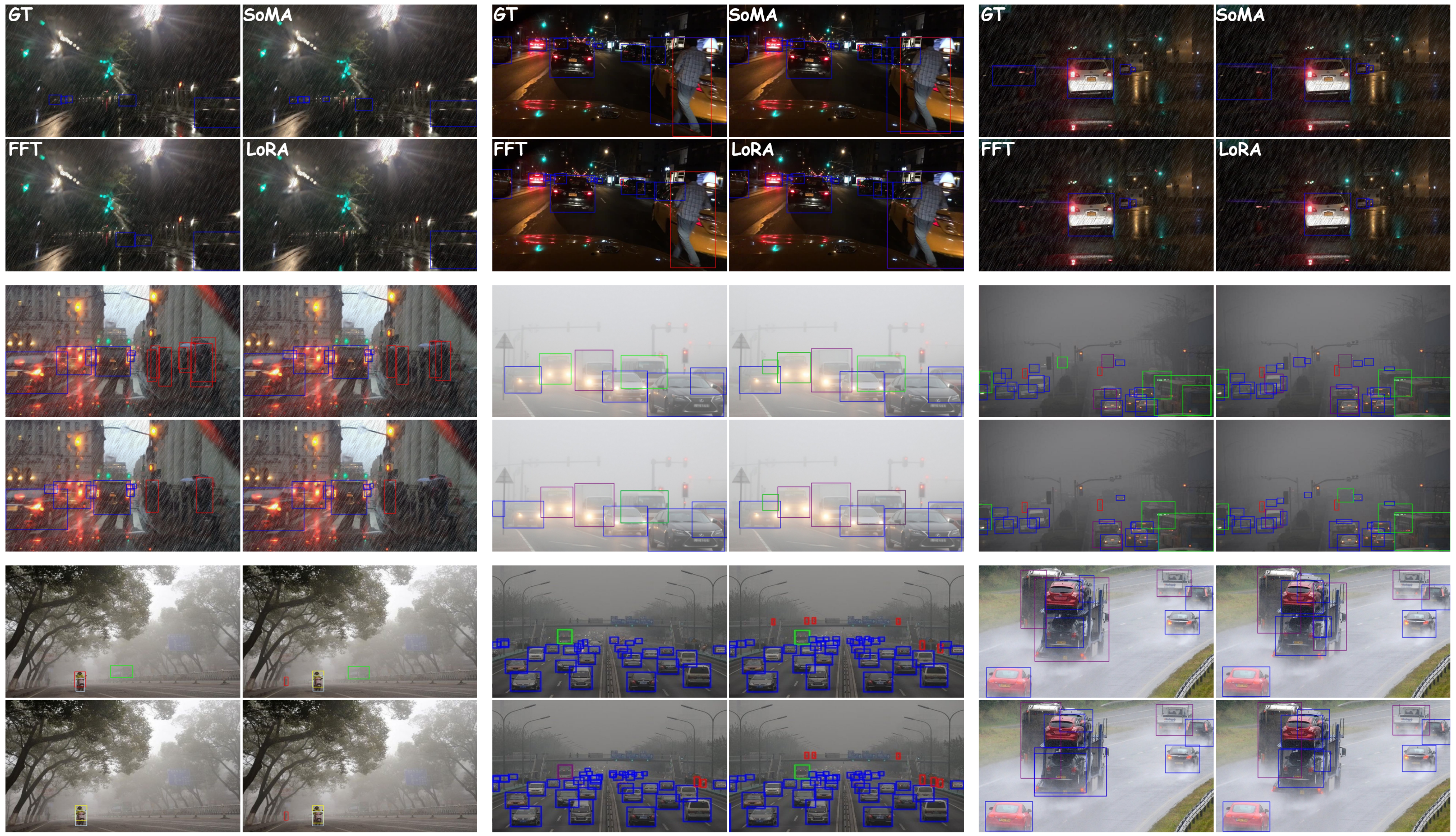}
    \vspace{-3.5mm}
    \caption{
    Detection results of SoMA on the adverse scene. The model is trained on Daytime-Sunny with DINOv2-L backbone.
    }
    \label{fig:supp_dgod_qualitative}
    \vspace{-3mm}
\end{figure*}

\noindent\textbf{DGSS and DGOD Qualitative Comparison.} 
Figures \ref{fig:supp_citys_qualitative}, \ref{fig:supp_bdd_qualitative}, and \ref{fig:supp_map_qualitative} depict DGSS prediction results on unseen domains for Cityscapes, BDD100k, and Mapillary, respectively, while Fig.~\ref{fig:supp_dgod_qualitative} provides detection results under various adverse conditions.
As evident from the visual comparisons above, SoMA demonstrates remarkable robustness to domain shifts resulting from diverse attributes (e.g., translucency, lighting conditions, road features, geographic variations, weather differences), while also excelling in fine-detail recognition compared to the selected baselines.

\begin{figure*}[h]
    \vspace{-1.5mm}
    \centering
    \includegraphics[width=\textwidth]{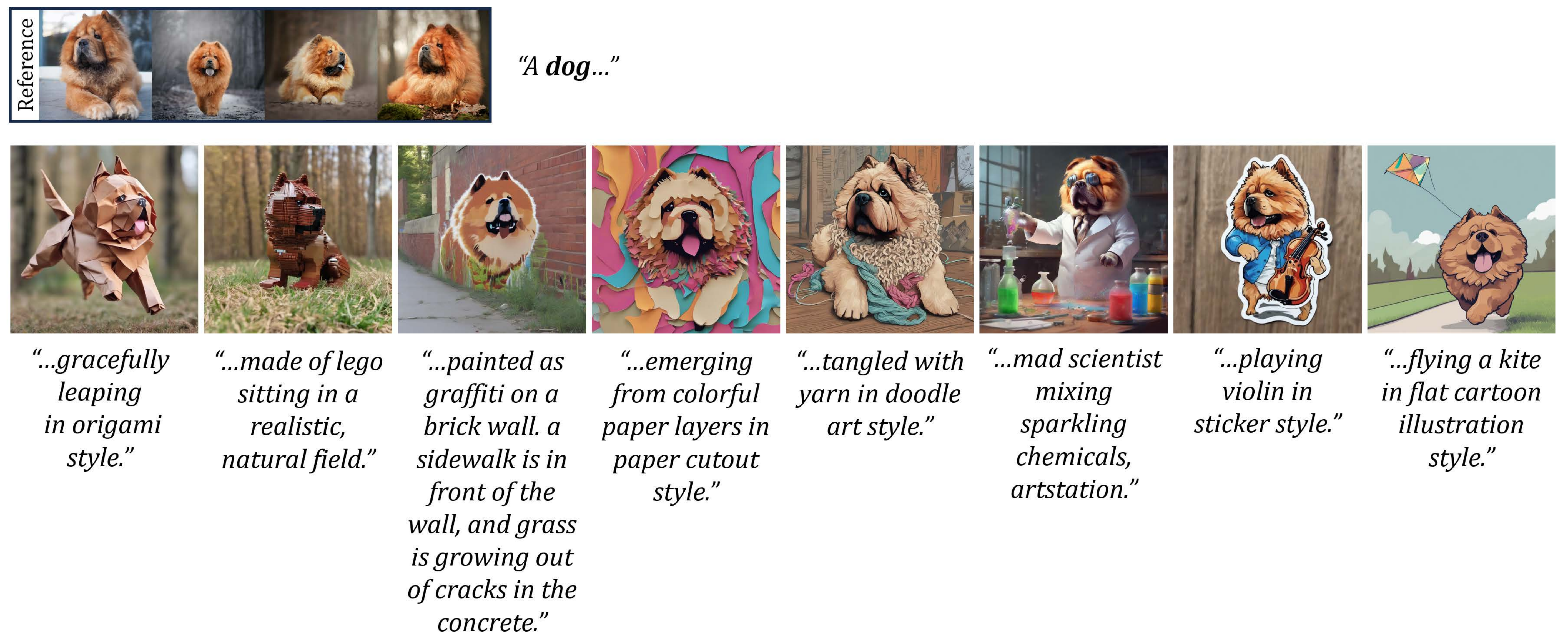}
    \vspace{-4.5mm}
    \caption{
    \textbf{Multiple subject-consistent synthesis results with prompts describing various domains.}
    SoMA effectively preserves SDXL's ability to generate images across diverse domains while learning new visual concepts. As a result, simply using prompts from multiple domains allows us to generate an image set of different domains that share the same subject.
    }
    \label{fig:supp_sdg1}
    \vspace{2mm}
\end{figure*}

\begin{figure*}[h]
    \vspace{-15mm}
    \centering
    \includegraphics[width=\textwidth]{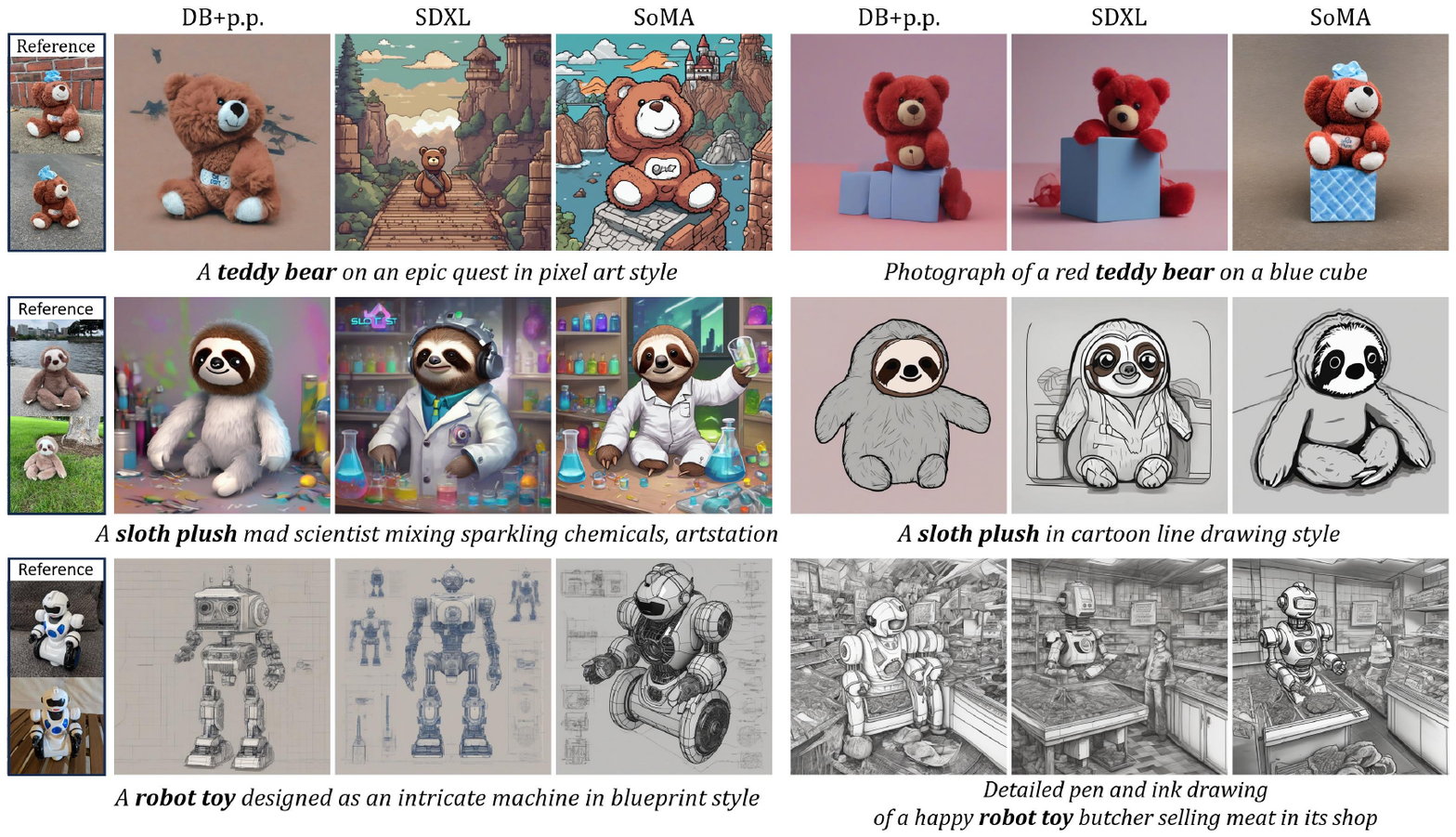}
    \vspace{-17mm}
    \caption{
    \textbf{Qualitative comparison to DreamBooth}~\cite{ruiz2023dreambooth} \textbf{with prior preservation (p.p.) loss.}
    }
    \label{fig:supp_sdg2}
    \vspace{-3mm}
\end{figure*}

\noindent\textbf{Subject Personalization.} Domain generalized recognition requires consistent processing of inputs from diverse domains, whereas domain generalized generation involves generating outputs across a range of domains.
Although large-scale Text-to-Image (T2I) models have convincingly demonstrated this ability, it can be compromised in subject personalization tasks involving fine-tuning.
As shown in Figures~\ref{fig:supp_sdg1} and \ref{fig:supp_sdg2}, integrating the SoMA framework in this case enables T2I models to fully leverage their generalization capability to synthesize target subjects in new domains.

\textbf{In summary, our proposed methods effectively facilitate domain-generalizable representation learning by maximally preserving pre-trained knowledge across diverse domains while learning task-specific features.}

\section{Discussion and Limitations}
Our adaptation approach introduces SVD as an interpretable tool applied to raw weight matrices, offering a fresh perspective on domain generalization.
Within this perspective, we focus on tuning the minor singular components to preserve the integrity of generalizable components with minimal interference.
However, achieving further performance improvements will require a more structured and nuanced design space.
Questions such as whether focusing solely on the lowest spectral space is optimal, or how to identify and adjust specific singular components for particular tasks, remain as avenues for future exploration.
Additionally, we plan to investigate design choices such as setting different ranks for each block or examining whether the low-rank matrices $A$ and $B$ play distinct roles, analyzing how these decisions influence generalization performance.
Extending these comprehensive analyses to other domains where foundation models are primarily employed, such as LLM benchmarks and audio applications, would also be an exciting direction for future work.


\begin{table}[h]
    \begin{minipage}{0.48\textwidth}
    \centering
        \tablestyle{1.0pt}{1.01}
        \begin{tabular}{lccccccc} 
\toprule
\multicolumn{3}{c}{\textbf{\emph{Synthetic-to-Real Generalization}}}&\multicolumn{4}{c}{Test Domains (mIoU in \%)} \\ \cmidrule(r){4-7}
Methods &Backbone &Head &$\rightarrow$\emph{Citys.} &$\rightarrow$\emph{BDD} &$\rightarrow$\emph{Map.} &Avg. \\\midrule
\multicolumn{7}{c}{\emph{Single-source DGSS Trained on \underline{GTAV}}} \\
\addlinespace[2pt]
\normal IBN-Net~\cite{ibnnet} &RN50 &DL-V3+ &33.85 &32.30 &37.75 &34.63 \\
\normal RobustNet~\cite{choi2021robustnet} &RN50 &DL-V3+ &36.58 &35.20 &40.33 &37.37 \\
\normal DRPC~\cite{randomization} &RN101 &FCN &42.53 &38.72 &38.05 &39.77 \\
\normal SiamDoGe~\cite{wu2022siamdoge} &RN50 &DL-V3+ &42.96 &37.54 &40.64 &40.38 \\
\normal DIRL~\cite{xu2022dirl} &RN50 &DL-V3+ &41.04 &39.15 &41.60 &40.60 \\
\normal GTR~\cite{peng2021globallocalTR} &RN101 &- &43.70 &39.60 &39.10 &40.80 \\
\normal AdvStyle~\cite{advstyle} &RN101 &DL-V3+ &43.44 &40.32 &41.96 &41.91 \\
\normal PintheMem~\cite{kim2022pin} &RN101 &DL-V2 & 44.90 &39.71 & 41.31 &41.97 \\
\normal MRFP+~\cite{udupa2024mrfp} &RN50 &DL-V3+ & 42.40 & 39.55 &44.93 &42.29 \\
\normal SAN-SAW~\cite{sansaw} &RN101 &DL-V3+ &45.33 &41.18 &40.77 &42.43 \\
\normal SPC~\cite{spc} &RN50 &DL-V3+ &44.10 &40.46 &45.51 &43.36 \\
\normal BlindNet~\cite{ahn2024blindnet} &RN50 &DL-V3+ &45.72 &41.32 &47.08 &44.71 \\
\normal WildNet~\cite{lee2022wildnet} &RN101 &DL-V3+ &45.79 &41.73 &47.08 &44.87 \\
\normal SHADE~\cite{shade} &RN101 &DL-V3+ &46.66 &43.66 &45.50 &45.27 \\
\normal PASTA~\cite{chattopadhyay2023pasta} &RN101 &DL-V3+ &45.33 &42.32 &48.60 &45.42 \\
\cellcolor[HTML]{efefef}\normal SoMA (Ours) &\cellcolor[HTML]{efefef}RN101 &\cellcolor[HTML]{efefef}M2F &\cellcolor[HTML]{efefef}41.23 &\cellcolor[HTML]{efefef}45.57 &\cellcolor[HTML]{efefef}49.71 &\cellcolor[HTML]{efefef}45.50 \\
\normal MoDify~\cite{modify} & RN101 &DL-V2 &48.80 &44.20 &47.50 & 46.80 \\
\normal TLDR~\cite{kim2023texture} &RN101 &DL-V3+ &47.58 &44.88 &48.80 &47.09 \\
\normal FAMix~\cite{famix} &CLIP RN101 &DL-V3+ &49.47 &46.40 &51.97 &49.28 \\
\normal CMFormer~\cite{cmformer} &Swin-L &- &55.31 &49.91 &60.09 &55.10 \\
\cellcolor[HTML]{efefef}\normal SoMA (Ours) &\cellcolor[HTML]{efefef}Swin-L &\cellcolor[HTML]{efefef}M2F &\cellcolor[HTML]{efefef}56.91 &\cellcolor[HTML]{efefef}51.98 &\cellcolor[HTML]{efefef}60.73 &\cellcolor[HTML]{efefef}56.54 \\
\normal DGInStyle~\cite{jia2025dginstyle} &MiT-B5 &HRDA &58.63 &52.25 &62.47 &57.78 \\
\normal DIDEX~\cite{diffusiondomainextension} &MiT-B5 &\scalebox{0.85}{DAFormer} &62.00 &54.30 &63.00 &59.70 \\
\normal CLOUDS~\cite{clouds} &CLIP CN-L &M2F &60.20 &57.40 &67.00 &61.50 \\
\normal VLTSeg~\cite{hummer2023vltseg} &EVA02-L &M2F &65.30 &58.30 &66.00 &63.20 \\
\normal Rein~\cite{rein} &EVA02-L &M2F &65.30 &60.50 &64.90 &63.60 \\
\normal FADA~\cite{bi2024fada} &EVA02-L &M2F &66.70 &61.90 &66.10 &64.90 \\ 
\normal tqdm~\cite{tqdm} &EVA02-L &M2F &68.88 &59.18 &70.10 &66.05 \\
\cellcolor[HTML]{efefef}\normal SoMA~(Ours) &\cellcolor[HTML]{efefef}EVA02-L &\cellcolor[HTML]{efefef}M2F &\cellcolor[HTML]{efefef}68.05 &\cellcolor[HTML]{efefef}60.81 &\cellcolor[HTML]{efefef}68.33 &\cellcolor[HTML]{efefef}65.73 \\
\cellcolor[HTML]{efefef}\tta SoMA~(Ours) &\cellcolor[HTML]{efefef}EVA02-L &\cellcolor[HTML]{efefef}M2F &\cellcolor[HTML]{efefef}69.94 &\cellcolor[HTML]{efefef}62.48&\cellcolor[HTML]{efefef}68.33 &\cellcolor[HTML]{efefef}66.92 \\
\peft DoRA~\cite{dora} &DINOv2-L &M2F & 66.12 &59.31 &67.07 & 64.17 \\
\peft VPT~\cite{jia2022vpt} &DINOv2-L &M2F & 68.75 &58.64 &68.32 & 65.24 \\
\normal SET~\cite{set} &DINOv2-L &M2F &68.06 &61.64 &67.68 &65.79 \\
\normal FADA~\cite{bi2024fada} &DINOv2-L &M2F &68.23 &61.94 &68.09 & 66.09 \\
\peft \scalebox{0.85}{AdaptFormer}~\cite{chen2022adaptformer} &DINOv2-L &M2F &70.10 & 59.81 & 68.77 & 66.23 \\
\peft SSF~\cite{ssf} &DINOv2-L &M2F & 68.97 &61.30 & 68.77 &66.35 \\
\peft LoRA~\cite{hu2022lora} &DINOv2-L &M2F & 70.13 &60.13 &70.42 & 66.89 \\
\normal $\text{Rein}^{\dagger}$~\cite{rein} &DINOv2-L &M2F &69.19 &60.01 &69.06 &66.09 \\
\tta $\text{Rein}^{\dagger}$~\cite{rein} &DINOv2-L &M2F &70.68 &62.51 &69.61 &67.60 \\
\cellcolor[HTML]{efefef}\normal SoMA~(Ours) &\cellcolor[HTML]{efefef}DINOv2-L &\cellcolor[HTML]{efefef}M2F &\cellcolor[HTML]{efefef}71.82 &\cellcolor[HTML]{efefef}61.31 &\cellcolor[HTML]{efefef}\textbf{71.67} &\cellcolor[HTML]{efefef}68.27 \\
\cellcolor[HTML]{efefef}\tta SoMA~(Ours) &\cellcolor[HTML]{efefef}DINOv2-L &\cellcolor[HTML]{efefef}M2F &\cellcolor[HTML]{efefef}\textbf{73.63} &\cellcolor[HTML]{efefef}\textbf{63.33}&\cellcolor[HTML]{efefef}70.98 &\cellcolor[HTML]{efefef}\textbf{69.31} \\
\midrule
\multicolumn{7}{c}{\emph{Multi-source DGSS Trained on \underline{GTAV + SYNTHIA}}} \\
\addlinespace[2pt]
\normal RobustNet~\cite{choi2021robustnet} &RN50 &DL-V3+ & 37.69 &34.09 &38.49 &36.76 \\
\normal AdvStyle~\cite{advstyle} &RN50 &DL-V3+ &39.29 &39.26 &41.14 &39.90 \\
\normal DIGA~\cite{shen2023diga} &RN101 &DL-V2 &46.43 &33.87 &43.51 &41.27 \\
\normal PintheMem~\cite{kim2022pin} &RN50 &DL-V3+ &44.51 &38.07 &42.70 &41.76 \\
\normal MRFP+~\cite{udupa2024mrfp} &RN50 &DL-V3+ & 46.18 & 41.13 &45.28 &44.24 \\
\normal SHADE~\cite{shade} &RN50 &DL-V3+ &47.43 &40.30 &47.60 &45.11 \\
\normal TLDR~\cite{kim2023texture} &RN50 &DL-V3+ &48.83 &42.58 &47.80 &46.40 \\
\normal SPC~\cite{spc} &RN101 &DL-V3+ &47.93 &43.62 &48.79 &46.78 \\
\normal FAMix~\cite{famix} &CLIP RN50 &DL-V3+ &49.41 &45.51 & 51.61 & 48.84 \\
\normal $\text{Rein}^{\dagger}$~\cite{rein} &DINOv2-L &M2F &72.17 &61.53 &70.69 &68.13 \\
\cellcolor[HTML]{efefef}\normal SoMA~(Ours) &\cellcolor[HTML]{efefef}DINOv2-L &\cellcolor[HTML]{efefef}M2F &\cellcolor[HTML]{efefef}73.16 &\cellcolor[HTML]{efefef}61.90 &\cellcolor[HTML]{efefef}72.73 &\cellcolor[HTML]{efefef}69.26 \\
\cellcolor[HTML]{efefef}\tta SoMA~(Ours) &\cellcolor[HTML]{efefef}DINOv2-L &\cellcolor[HTML]{efefef}M2F &\cellcolor[HTML]{efefef}\textbf{74.85} &\cellcolor[HTML]{efefef}\textbf{63.59}&\cellcolor[HTML]{efefef}\textbf{73.92} &\cellcolor[HTML]{efefef}\textbf{70.79} \\
\bottomrule
\end{tabular}
        \caption{
        Comparison of the proposed SoMA with existing DGSS \normal and PEFT \peft methods under various \textbf{synthetic-to-real settings}.
        }\label{tab: supp_full_dgss_s2r}
    \end{minipage}
\end{table}

\begin{table}[h]
    \begin{minipage}{0.48\textwidth}
    \centering
        \tablestyle{3.2pt}{1.0}
        \begin{tabular}{lccccc}  
\toprule
\multicolumn{3}{c}{\textbf{\emph{Real-to-Real Generalization}}}&\multicolumn{3}{c}{Test Domains (mIoU in \%)} \\ \cmidrule(r){4-6}
Methods &Backbone &Head &$\rightarrow$\emph{BDD} &$\rightarrow$\emph{Map.} &Avg. \\\midrule
\multicolumn{6}{c}{\emph{Single-source DGSS Trained on \underline{Cityscapes}}} \\
\addlinespace[2pt]
\normal RobustNet~\cite{choi2021robustnet} &RN50 &DL-V3+ &50.73 &58.64 &54.69 \\
\normal WildNet~\cite{lee2022wildnet} &RN50 &DL-V3+ &50.94 &58.79 &54.87 \\
\normal SiamDoGe~\cite{wu2022siamdoge} &RN50 &DL-V3+ &51.53 &59.00 &55.27 \\
\normal SHADE~\cite{shade} &RN50 &DL-V3+ &50.95 &60.67 &55.81 \\
\normal BlindNet~\cite{ahn2024blindnet} &RN50 &DL-V3+ &51.84 &60.18 &56.01 \\
\normal FAMix~\cite{famix} &CLIP RN50 &DL-V3+ & 54.07 &58.72 &56.40 \\
\normal SAN-SAW~\cite{sansaw} &RN101 &DL-V3+ &54.73 &61.27 &58.00 \\
\normal HGFormer~\cite{ding2023hgformer} &Swin-L &- &61.50 &72.10 &66.80   \\
\normal CMFormer~\cite{cmformer} &Swin-L &- &62.60 &73.60 &68.10 \\
\normal tqdm~\cite{tqdm} &EVA02-L &M2F &64.72 &76.15 &70.44 \\
\normal FADA~\cite{bi2024fada} &DINOv2-L &M2F &65.12 &75.86 &70.49 \\
\normal $\text{Rein}^{\dagger}$~\cite{rein} &DINOv2-L &M2F &66.53 &75.18 &70.86 \\
\cellcolor[HTML]{efefef}\normal SoMA~(Ours) &\cellcolor[HTML]{efefef}DINOv2-L & \cellcolor[HTML]{efefef}M2F &\cellcolor[HTML]{efefef}67.02 &\cellcolor[HTML]{efefef}76.45 &\cellcolor[HTML]{efefef}71.74 \\
\cellcolor[HTML]{efefef}\tta SoMA~(Ours) &\cellcolor[HTML]{efefef}DINOv2-L &\cellcolor[HTML]{efefef}M2F&\cellcolor[HTML]{efefef}\textbf{68.08} &\cellcolor[HTML]{efefef}\textbf{77.87}&\cellcolor[HTML]{efefef}\textbf{72.98} \\
\bottomrule
\end{tabular}
        \caption{
        \textbf{Real-to-real DGSS comparison.}
        }\label{tab: supp_full_dgss_r2r}
        \vspace{3mm}
    \end{minipage} \\
    \begin{minipage}{0.48\textwidth}
    \centering
        \tablestyle{4.2pt}{1.0}
        \begin{tabular}{lccccc}  
\toprule
\textbf{\emph{\scalebox{0.85}{Clear-to-Adverse Weather}}} &\multicolumn{5}{c}{ACDC~\cite{acdc} Test Domains (mIoU in \%)} \\ \cmidrule(lr){2-6}
Methods &$\rightarrow$\emph{Night} &$\rightarrow$\emph{Snow} &$\rightarrow$\emph{Fog} &$\rightarrow$\emph{Rain} &Avg. \\\midrule
\multicolumn{6}{c}{\emph{Single-source DGSS Trained on \underline{Cityscapes}}} \\
\addlinespace[2pt]
\normal CMFormer~\cite{cmformer} &33.7 &64.3 &77.8 &67.6 &60.9   \\
\normal SET~\cite{set} &57.3 &73.7 &80.1 &74.8 &71.5 \\
\normal FADA~\cite{bi2024fada} &\textbf{57.4} &73.5 &80.2 &75.0 &71.5 \\
\cellcolor[HTML]{efefef}\normal SoMA (Ours) &\cellcolor[HTML]{efefef}52.4 &\cellcolor[HTML]{efefef}\textbf{74.6} &\cellcolor[HTML]{efefef}\textbf{84.1} &\cellcolor[HTML]{efefef}\textbf{75.5} &\cellcolor[HTML]{efefef}\textbf{71.7} \\
\bottomrule
\end{tabular}
        \caption{
        Results on \textbf{Cityscapes $\rightarrow$ ACDC \underline{validation} set.}
        }\label{tab: supp_dgss_acdc_val}
        \vspace{1mm}
    \end{minipage}
\end{table}

\newpage

{
\small
\bibliographystyle{ieeenat_fullname}
\bibliography{main}

\begin{thebibliography}{93}
\providecommand{\natexlab}[1]{#1}
\providecommand{\url}[1]{\texttt{#1}}
\expandafter\ifx\csname urlstyle\endcsname\relax
  \providecommand{\doi}[1]{doi: #1}\else
  \providecommand{\doi}{doi: \begingroup \urlstyle{rm}\Url}\fi

\bibitem[Ahn et~al.(2024)Ahn, Yang, Choi, and Lim]{ahn2024blindnet}
Woo-Jin Ahn, Geun-Yeong Yang, Hyun-Duck Choi, and Myo-Taeg Lim.
\newblock Style blind domain generalized semantic segmentation via covariance alignment and semantic consistence contrastive learning.
\newblock In \emph{CVPR}, 2024.

\bibitem[Awais et~al.(2023)Awais, Naseer, Khan, Anwer, Cholakkal, Shah, Yang, and Khan]{awais2023foundational}
Muhammad Awais, Muzammal Naseer, Salman Khan, Rao~Muhammad Anwer, Hisham Cholakkal, Mubarak Shah, Ming-Hsuan Yang, and Fahad~Shahbaz Khan.
\newblock Foundational models defining a new era in vision: A survey and outlook.
\newblock \emph{arXiv:2307.13721}, 2023.

\bibitem[Benigmim et~al.(2024)Benigmim, Roy, Essid, Kalogeiton, and Lathuili\`ere]{clouds}
Yasser Benigmim, Subhankar Roy, Slim Essid, Vicky Kalogeiton, and St\'ephane Lathuili\`ere.
\newblock Collaborating foundation models for domain generalized semantic segmentation.
\newblock In \emph{CVPR}, 2024.

\bibitem[Bi et~al.(2024{\natexlab{a}})Bi, Yi, Zheng, Zhan, Huang, Ji, Li, and Zheng]{bi2024fada}
Qi Bi, Jingjun Yi, Hao Zheng, Haolan Zhan, Yawen Huang, Wei Ji, Yuexiang Li, and Yefeng Zheng.
\newblock Learning frequency-adapted vision foundation model for domain generalized semantic segmentation.
\newblock In \emph{NeurIPS}, 2024{\natexlab{a}}.

\bibitem[Bi et~al.(2024{\natexlab{b}})Bi, You, and Gevers]{cmformer}
Qi Bi, Shaodi You, and Theo Gevers.
\newblock Learning content-enhanced mask transformer for domain generalized urban-scene segmentation.
\newblock In \emph{AAAI}, 2024{\natexlab{b}}.

\bibitem[Biderman et~al.(2024)Biderman, Ortiz, Portes, Paul, Greengard, Jennings, King, Havens, Chiley, Frankle, et~al.]{biderman2024loraforgetless}
Dan Biderman, Jose~Gonzalez Ortiz, Jacob Portes, Mansheej Paul, Philip Greengard, Connor Jennings, Daniel King, Sam Havens, Vitaliy Chiley, Jonathan Frankle, et~al.
\newblock Lora learns less and forgets less.
\newblock \emph{TMLR}, 2024.

\bibitem[Carion et~al.(2020)Carion, Massa, Synnaeve, Usunier, Kirillov, and Zagoruyko]{detr}
Nicolas Carion, Francisco Massa, Gabriel Synnaeve, Nicolas Usunier, Alexander Kirillov, and Sergey Zagoruyko.
\newblock End-to-end object detection with transformers.
\newblock In \emph{ECCV}, 2020.

\bibitem[Caron et~al.(2021)Caron, Touvron, Misra, J{\'e}gou, Mairal, Bojanowski, and Joulin]{dino}
Mathilde Caron, Hugo Touvron, Ishan Misra, Herv{\'e} J{\'e}gou, Julien Mairal, Piotr Bojanowski, and Armand Joulin.
\newblock Emerging properties in self-supervised vision transformers.
\newblock In \emph{ICCV}, 2021.

\bibitem[Chang et~al.(2024)Chang, Lee, Kim, Kim, Lee, Ji, Jang, and Kim]{chang2024unified}
Gyusam Chang, Jiwon Lee, Donghyun Kim, Jinkyu Kim, Dongwook Lee, Daehyun Ji, Sujin Jang, and Sangpil Kim.
\newblock Unified domain generalization and adaptation for multi-view 3d object detection.
\newblock In \emph{NeurIPS}, 2024.

\bibitem[Chattopadhyay et~al.(2023)Chattopadhyay, Sarangmath, Vijaykumar, and Hoffman]{chattopadhyay2023pasta}
Prithvijit Chattopadhyay, Kartik Sarangmath, Vivek Vijaykumar, and Judy Hoffman.
\newblock Pasta: Proportional amplitude spectrum training augmentation for syn-to-real domain generalization.
\newblock In \emph{ICCV}, 2023.

\bibitem[Chen et~al.(2019)Chen, Wang, Pang, Cao, Xiong, Li, Sun, Feng, Liu, Xu, Zhang, Cheng, Zhu, Cheng, Zhao, Li, Lu, Zhu, Wu, Dai, Wang, Shi, Ouyang, Loy, and Lin]{mmdetection}
Kai Chen, Jiaqi Wang, Jiangmiao Pang, Yuhang Cao, Yu Xiong, Xiaoxiao Li, Shuyang Sun, Wansen Feng, Ziwei Liu, Jiarui Xu, Zheng Zhang, Dazhi Cheng, Chenchen Zhu, Tianheng Cheng, Qijie Zhao, Buyu Li, Xin Lu, Rui Zhu, Yue Wu, Jifeng Dai, Jingdong Wang, Jianping Shi, Wanli Ouyang, Chen~Change Loy, and Dahua Lin.
\newblock {MMDetection}: Open mmlab detection toolbox and benchmark.
\newblock \emph{arXiv:1906.07155}, 2019.

\bibitem[Chen et~al.(2022)Chen, Ge, Tong, Wang, Song, Wang, and Luo]{chen2022adaptformer}
Shoufa Chen, Chongjian Ge, Zhan Tong, Jiangliu Wang, Yibing Song, Jue Wang, and Ping Luo.
\newblock Adaptformer: Adapting vision transformers for scalable visual recognition.
\newblock In \emph{NeurIPS}, 2022.

\bibitem[Cheng et~al.(2022)Cheng, Misra, Schwing, Kirillov, and Girdhar]{cheng2022mask2former}
Bowen Cheng, Ishan Misra, Alexander~G Schwing, Alexander Kirillov, and Rohit Girdhar.
\newblock Masked-attention mask transformer for universal image segmentation.
\newblock In \emph{CVPR}, 2022.

\bibitem[Choi et~al.(2021)Choi, Jung, Yun, Kim, Kim, and Choo]{choi2021robustnet}
Sungha Choi, Sanghun Jung, Huiwon Yun, Joanne~T Kim, Seungryong Kim, and Jaegul Choo.
\newblock Robustnet: Improving domain generalization in urban-scene segmentation via instance selective whitening.
\newblock In \emph{CVPR}, 2021.

\bibitem[Contributors(2020)]{mmseg2020}
MMSegmentation Contributors.
\newblock {MMSegmentation}: Openmmlab semantic segmentation toolbox and benchmark.
\newblock \url{https://github.com/open-mmlab/mmsegmentation}, 2020.

\bibitem[Cordts et~al.(2016)Cordts, Omran, Ramos, Rehfeld, Enzweiler, Benenson, Franke, Roth, and Schiele]{citys}
Marius Cordts, Mohamed Omran, Sebastian Ramos, Timo Rehfeld, Markus Enzweiler, Rodrigo Benenson, Uwe Franke, Stefan Roth, and Bernt Schiele.
\newblock The cityscapes dataset for semantic urban scene understanding.
\newblock In \emph{CVPR}, 2016.

\bibitem[Danish et~al.(2024)Danish, Khan, Munir, Sarfraz, and Ali]{divalign}
Muhammad~Sohail Danish, Muhammad~Haris Khan, Muhammad~Akhtar Munir, M~Saquib Sarfraz, and Mohsen Ali.
\newblock Improving single domain-generalized object detection: A focus on diversification and alignment.
\newblock In \emph{CVPR}, 2024.

\bibitem[Deng et~al.(2009)Deng, Dong, Socher, Li, Li, and Fei-Fei]{deng2009imagenet}
Jia Deng, Wei Dong, Richard Socher, Li-Jia Li, Kai Li, and Li Fei-Fei.
\newblock Imagenet: A large-scale hierarchical image database.
\newblock In \emph{CVPR}, 2009.

\bibitem[Ding et~al.(2023)Ding, Xue, Xia, Schiele, and Dai]{ding2023hgformer}
Jian Ding, Nan Xue, Gui-Song Xia, Bernt Schiele, and Dengxin Dai.
\newblock Hgformer: Hierarchical grouping transformer for domain generalized semantic segmentation.
\newblock In \emph{CVPR}, 2023.

\bibitem[Dosovitskiy et~al.(2021)Dosovitskiy, Beyer, Kolesnikov, Weissenborn, Zhai, Unterthiner, Dehghani, Minderer, Heigold, Gelly, Uszkoreit, and Houlsby]{vit}
Alexey Dosovitskiy, Lucas Beyer, Alexander Kolesnikov, Dirk Weissenborn, Xiaohua Zhai, Thomas Unterthiner, Mostafa Dehghani, Matthias Minderer, Georg Heigold, Sylvain Gelly, Jakob Uszkoreit, and Neil Houlsby.
\newblock An image is worth 16x16 words: Transformers for image recognition at scale.
\newblock In \emph{ICLR}, 2021.

\bibitem[Eckart and Young(1936)]{svd}
Carl Eckart and Gale Young.
\newblock The approximation of one matrix by another of lower rank.
\newblock \emph{Psychometrika}, 1936.

\bibitem[Fahes et~al.(2024)Fahes, Vu, Bursuc, P{\'e}rez, and de~Charette]{famix}
Mohammad Fahes, Tuan-Hung Vu, Andrei Bursuc, Patrick P{\'e}rez, and Raoul de Charette.
\newblock A simple recipe for language-guided domain generalized segmentation.
\newblock In \emph{CVPR}, 2024.

\bibitem[Fan et~al.(2023)Fan, Segu, Tai, Yu, Tang, Schiele, and Dai]{NP}
Qi Fan, Mattia Segu, Yu-Wing Tai, Fisher Yu, Chi-Keung Tang, Bernt Schiele, and Dengxin Dai.
\newblock Towards robust object detection invariant to real-world domain shifts.
\newblock In \emph{ICLR}, 2023.

\bibitem[Fang et~al.(2024)Fang, Sun, Wang, Huang, Wang, and Cao]{eva02}
Yuxin Fang, Quan Sun, Xinggang Wang, Tiejun Huang, Xinlong Wang, and Yue Cao.
\newblock Eva-02: A visual representation for neon genesis.
\newblock \emph{IVC}, 2024.

\bibitem[Frenkel et~al.(2024)Frenkel, Vinker, Shamir, and Cohen-Or]{frenkel2025b-lora}
Yarden Frenkel, Yael Vinker, Ariel Shamir, and Daniel Cohen-Or.
\newblock Implicit style-content separation using b-lora.
\newblock In \emph{ECCV}, 2024.

\bibitem[Golatkar et~al.(2019)Golatkar, Achille, and Soatto]{time_matters}
Aditya~Sharad Golatkar, Alessandro Achille, and Stefano Soatto.
\newblock Time matters in regularizing deep networks: Weight decay and data augmentation affect early learning dynamics, matter little near convergence.
\newblock In \emph{NeurIPS}, 2019.

\bibitem[G{\'o}mez et~al.(2023)G{\'o}mez, Silva, Seoane, Borr{\'a}s, Noriega, Ros, Iglesias-Guitian, and L{\'o}pez]{urbansyn}
Jose~L G{\'o}mez, Manuel Silva, Antonio Seoane, Agn{\`e}s Borr{\'a}s, Mario Noriega, Germ{\'a}n Ros, Jose~A Iglesias-Guitian, and Antonio~M L{\'o}pez.
\newblock All for one, and one for all: Urbansyn dataset, the third musketeer of synthetic driving scenes.
\newblock \emph{arXiv:2312.12176}, 2023.

\bibitem[He et~al.(2022{\natexlab{a}})He, Zhou, Ma, Berg-Kirkpatrick, and Neubig]{unifiedviewadapter}
Junxian He, Chunting Zhou, Xuezhe Ma, Taylor Berg-Kirkpatrick, and Graham Neubig.
\newblock Towards a unified view of parameter-efficient transfer learning.
\newblock In \emph{ICLR}, 2022{\natexlab{a}}.

\bibitem[He et~al.(2015)He, Zhang, Ren, and Sun]{heinit}
Kaiming He, Xiangyu Zhang, Shaoqing Ren, and Jian Sun.
\newblock Delving deep into rectifiers: Surpassing human-level performance on imagenet classification.
\newblock In \emph{ICCV}, 2015.

\bibitem[He et~al.(2016)He, Zhang, Ren, and Sun]{resnet}
Kaiming He, Xiangyu Zhang, Shaoqing Ren, and Jian Sun.
\newblock Deep residual learning for image recognition.
\newblock In \emph{CVPR}, 2016.

\bibitem[He et~al.(2022{\natexlab{b}})He, Chen, Xie, Li, Doll{\'a}r, and Girshick]{he2022masked}
Kaiming He, Xinlei Chen, Saining Xie, Yanghao Li, Piotr Doll{\'a}r, and Ross Girshick.
\newblock Masked autoencoders are scalable vision learners.
\newblock In \emph{CVPR}, 2022{\natexlab{b}}.

\bibitem[Hoyer et~al.(2022)Hoyer, Dai, and Van~Gool]{hoyer2022daformer}
Lukas Hoyer, Dengxin Dai, and Luc Van~Gool.
\newblock Daformer: Improving network architectures and training strategies for domain-adaptive semantic segmentation.
\newblock In \emph{CVPR}, 2022.

\bibitem[Hu et~al.(2022)Hu, yelong shen, Wallis, Allen-Zhu, Li, Wang, Wang, and Chen]{hu2022lora}
Edward~J Hu, yelong shen, Phillip Wallis, Zeyuan Allen-Zhu, Yuanzhi Li, Shean Wang, Lu Wang, and Weizhu Chen.
\newblock Lo{RA}: Low-rank adaptation of large language models.
\newblock In \emph{ICLR}, 2022.

\bibitem[Hu et~al.(2024)Hu, Zhang, Qi, Shi, and Gao]{pego}
Jiajun Hu, Jian Zhang, Lei Qi, Yinghuan Shi, and Yang Gao.
\newblock Learn to preserve and diversify: Parameter-efficient group with orthogonal regularization for domain generalization.
\newblock In \emph{ECCV}, 2024.

\bibitem[Huang et~al.(2023)Huang, Chen, Li, Li, Li, Song, Yan, and Xiong]{spc}
Wei Huang, Chang Chen, Yong Li, Jiacheng Li, Cheng Li, Fenglong Song, Youliang Yan, and Zhiwei Xiong.
\newblock Style projected clustering for domain generalized semantic segmentation.
\newblock In \emph{CVPR}, 2023.

\bibitem[H{\"u}mmer et~al.(2024)H{\"u}mmer, Schwonberg, Zhong, Cao, Knoll, and Gottschalk]{hummer2023vltseg}
Christoph H{\"u}mmer, Manuel Schwonberg, Liangwei Zhong, Hu Cao, Alois Knoll, and Hanno Gottschalk.
\newblock Vltseg: Simple transfer of clip-based vision-language representations for domain generalized semantic segmentation.
\newblock In \emph{ACCV}, 2024.

\bibitem[Jia et~al.(2022)Jia, Tang, Chen, Cardie, Belongie, Hariharan, and Lim]{jia2022vpt}
Menglin Jia, Luming Tang, Bor-Chun Chen, Claire Cardie, Serge Belongie, Bharath Hariharan, and Ser-Nam Lim.
\newblock Visual prompt tuning.
\newblock In \emph{ECCV}, 2022.

\bibitem[Jia et~al.(2024)Jia, Hoyer, Huang, Wang, Van~Gool, Schindler, and Obukhov]{jia2025dginstyle}
Yuru Jia, Lukas Hoyer, Shengyu Huang, Tianfu Wang, Luc Van~Gool, Konrad Schindler, and Anton Obukhov.
\newblock Dginstyle: Domain-generalizable semantic segmentation with image diffusion models and stylized semantic control.
\newblock In \emph{ECCV}, 2024.

\bibitem[Jiang et~al.(2023)Jiang, Huang, Jin, and Lu]{modify}
Xueying Jiang, Jiaxing Huang, Sheng Jin, and Shijian Lu.
\newblock Domain generalization via balancing training difficulty and model capability.
\newblock In \emph{ICCV}, 2023.

\bibitem[Jing et~al.(2023)Jing, Zhen, Li, and Snoek]{jing2023order}
Mengmeng Jing, Xiantong Zhen, Jingjing Li, and Cees~GM Snoek.
\newblock Order-preserving consistency regularization for domain adaptation and generalization.
\newblock In \emph{ICCV}, 2023.

\bibitem[Kamann and Rother(2020)]{paintingbynumbers}
Christoph Kamann and Carsten Rother.
\newblock Increasing the robustness of semantic segmentation models with painting-by-numbers.
\newblock In \emph{ECCV}, 2020.

\bibitem[Kim et~al.(2022)Kim, Lee, Park, Min, and Sohn]{kim2022pin}
Jin Kim, Jiyoung Lee, Jungin Park, Dongbo Min, and Kwanghoon Sohn.
\newblock Pin the memory: Learning to generalize semantic segmentation.
\newblock In \emph{CVPR}, 2022.

\bibitem[Kim et~al.(2023)Kim, Kim, and Kim]{kim2023texture}
Sunghwan Kim, Dae-hwan Kim, and Hoseong Kim.
\newblock Texture learning domain randomization for domain generalized segmentation.
\newblock In \emph{ICCV}, 2023.

\bibitem[Kirillov et~al.(2019)Kirillov, Girshick, He, and Doll{\'a}r]{semfpn}
Alexander Kirillov, Ross Girshick, Kaiming He, and Piotr Doll{\'a}r.
\newblock Panoptic feature pyramid networks.
\newblock In \emph{CVPR}, 2019.

\bibitem[Lee et~al.(2022{\natexlab{a}})Lee, Seong, Lee, and Kim]{lee2022wildnet}
Suhyeon Lee, Hongje Seong, Seongwon Lee, and Euntai Kim.
\newblock Wildnet: Learning domain generalized semantic segmentation from the wild.
\newblock In \emph{CVPR}, 2022{\natexlab{a}}.

\bibitem[Lee et~al.(2022{\natexlab{b}})Lee, Son, and Kwak]{lee2022fifo}
Sohyun Lee, Taeyoung Son, and Suha Kwak.
\newblock Fifo: Learning fog-invariant features for foggy scene segmentation.
\newblock In \emph{CVPR}, 2022{\natexlab{b}}.

\bibitem[Lee et~al.(2024)Lee, Hong, Lim, and Myung]{oamix}
Wooju Lee, Dasol Hong, Hyungtae Lim, and Hyun Myung.
\newblock Object-aware domain generalization for object detection.
\newblock In \emph{AAAI}, 2024.

\bibitem[Lee et~al.(2023)Lee, Chen, Tajwar, Kumar, Yao, Liang, and Finn]{lee2023surgical}
Yoonho Lee, Annie~S Chen, Fahim Tajwar, Ananya Kumar, Huaxiu Yao, Percy Liang, and Chelsea Finn.
\newblock Surgical fine-tuning improves adaptation to distribution shifts.
\newblock In \emph{ICLR}, 2023.

\bibitem[Lester et~al.(2021)Lester, Al-Rfou, and Constant]{lester-etal-2021-power}
Brian Lester, Rami Al-Rfou, and Noah Constant.
\newblock The power of scale for parameter-efficient prompt tuning.
\newblock In \emph{EMNLP}, 2021.

\bibitem[Li et~al.(2024)Li, Wu, Wang, and Han]{pdod}
Deng Li, Aming Wu, Yaowei Wang, and Yahong Han.
\newblock Prompt-driven dynamic object-centric learning for single domain generalization.
\newblock In \emph{CVPR}, 2024.

\bibitem[Lian et~al.(2022)Lian, Zhou, Feng, and Wang]{ssf}
Dongze Lian, Daquan Zhou, Jiashi Feng, and Xinchao Wang.
\newblock Scaling \& shifting your features: A new baseline for efficient model tuning.
\newblock In \emph{NeurIPS}, 2022.

\bibitem[Liu et~al.(2024{\natexlab{a}})Liu, Wang, Yin, Molchanov, Wang, Cheng, and Chen]{dora}
Shih-Yang Liu, Chien-Yi Wang, Hongxu Yin, Pavlo Molchanov, Yu-Chiang~Frank Wang, Kwang-Ting Cheng, and Min-Hung Chen.
\newblock {D}o{RA}: Weight-decomposed low-rank adaptation.
\newblock In \emph{ICML}, 2024{\natexlab{a}}.

\bibitem[Liu et~al.(2024{\natexlab{b}})Liu, Zhou, Liu, Hao, Fan, and Tian]{liu2024unbiased}
Yajing Liu, Shijun Zhou, Xiyao Liu, Chunhui Hao, Baojie Fan, and Jiandong Tian.
\newblock Unbiased faster r-cnn for single-source domain generalized object detection.
\newblock In \emph{CVPR}, 2024{\natexlab{b}}.

\bibitem[Liu et~al.(2021)Liu, Lin, Cao, Hu, Wei, Zhang, Lin, and Guo]{liu2021swin}
Ze Liu, Yutong Lin, Yue Cao, Han Hu, Yixuan Wei, Zheng Zhang, Stephen Lin, and Baining Guo.
\newblock Swin transformer: Hierarchical vision transformer using shifted windows.
\newblock In \emph{ICCV}, 2021.

\bibitem[Meng et~al.(2024)Meng, Wang, and Zhang]{meng2024pissa}
Fanxu Meng, Zhaohui Wang, and Muhan Zhang.
\newblock Pissa: Principal singular values and singular vectors adaptation of large language models.
\newblock In \emph{NeurIPS}, 2024.

\bibitem[Neuhold et~al.(2017)Neuhold, Ollmann, Rota~Bulo, and Kontschieder]{map}
Gerhard Neuhold, Tobias Ollmann, Samuel Rota~Bulo, and Peter Kontschieder.
\newblock The mapillary vistas dataset for semantic understanding of street scenes.
\newblock In \emph{ICCV}, 2017.

\bibitem[Niemeijer et~al.(2024)Niemeijer, Schwonberg, Term{\"o}hlen, Schmidt, and Fingscheidt]{diffusiondomainextension}
Joshua Niemeijer, Manuel Schwonberg, Jan-Aike Term{\"o}hlen, Nico~M Schmidt, and Tim Fingscheidt.
\newblock Generalization by adaptation: Diffusion-based domain extension for domain-generalized semantic segmentation.
\newblock In \emph{WACV}, 2024.

\bibitem[Oquab et~al.(2023)Oquab, Darcet, Moutakanni, Vo, Szafraniec, Khalidov, Fernandez, Haziza, Massa, El-Nouby, et~al.]{oquab2023dinov2}
Maxime Oquab, Timoth{\'e}e Darcet, Th{\'e}o Moutakanni, Huy Vo, Marc Szafraniec, Vasil Khalidov, Pierre Fernandez, Daniel Haziza, Francisco Massa, Alaaeldin El-Nouby, et~al.
\newblock Dinov2: Learning robust visual features without supervision.
\newblock \emph{TMLR}, 2023.

\bibitem[Pak et~al.(2024)Pak, Woo, Kim, Kim, and Kim]{tqdm}
Byeonghyun Pak, Byeongju Woo, Sunghwan Kim, Dae-hwan Kim, and Hoseong Kim.
\newblock Textual query-driven mask transformer for domain generalized segmentation.
\newblock In \emph{ECCV}, 2024.

\bibitem[Pan et~al.(2018)Pan, Luo, Shi, and Tang]{ibnnet}
Xingang Pan, Ping Luo, Jianping Shi, and Xiaoou Tang.
\newblock Two at once: Enhancing learning and generalization capacities via ibn-net.
\newblock In \emph{ECCV}, 2018.

\bibitem[Pan et~al.(2024)Pan, Sun, Luo, Zhang, and Zhang]{nightseg}
Yuwen Pan, Rui Sun, Naisong Luo, Tianzhu Zhang, and Yongdong Zhang.
\newblock Exploring reliable matching with phase enhancement for night-time semantic segmentation.
\newblock In \emph{ECCV}, 2024.

\bibitem[Peng et~al.(2021)Peng, Lei, Liu, Zhang, and Liu]{peng2021globallocalTR}
Duo Peng, Yinjie Lei, Lingqiao Liu, Pingping Zhang, and Jun Liu.
\newblock Global and local texture randomization for synthetic-to-real semantic segmentation.
\newblock \emph{TIP}, 2021.

\bibitem[Peng et~al.(2022)Peng, Lei, Hayat, Guo, and Li]{sansaw}
Duo Peng, Yinjie Lei, Munawar Hayat, Yulan Guo, and Wen Li.
\newblock Semantic-aware domain generalized segmentation.
\newblock In \emph{CVPR}, 2022.

\bibitem[Podell et~al.(2024)Podell, English, Lacey, Blattmann, Dockhorn, M{\"u}ller, Penna, and Rombach]{podell2024sdxl}
Dustin Podell, Zion English, Kyle Lacey, Andreas Blattmann, Tim Dockhorn, Jonas M{\"u}ller, Joe Penna, and Robin Rombach.
\newblock {SDXL}: Improving latent diffusion models for high-resolution image synthesis.
\newblock In \emph{ICLR}, 2024.

\bibitem[Radford et~al.(2021)Radford, Kim, Hallacy, Ramesh, Goh, Agarwal, Sastry, Askell, Mishkin, Clark, et~al.]{radfordclip}
Alec Radford, Jong~Wook Kim, Chris Hallacy, Aditya Ramesh, Gabriel Goh, Sandhini Agarwal, Girish Sastry, Amanda Askell, Pamela Mishkin, Jack Clark, et~al.
\newblock Learning transferable visual models from natural language supervision.
\newblock In \emph{ICML}, 2021.

\bibitem[Ren et~al.(2015)Ren, He, Girshick, and Sun]{ren2015fasterrcnn}
Shaoqing Ren, Kaiming He, Ross Girshick, and Jian Sun.
\newblock Faster r-cnn: Towards real-time object detection with region proposal networks.
\newblock In \emph{NeurIPS}, 2015.

\bibitem[Richter et~al.(2016)Richter, Vineet, Roth, and Koltun]{gtav}
Stephan~R Richter, Vibhav Vineet, Stefan Roth, and Vladlen Koltun.
\newblock Playing for data: Ground truth from computer games.
\newblock In \emph{ECCV}, 2016.

\bibitem[Rombach et~al.(2022)Rombach, Blattmann, Lorenz, Esser, and Ommer]{LDM}
Robin Rombach, Andreas Blattmann, Dominik Lorenz, Patrick Esser, and Bj{\"o}rn Ommer.
\newblock High-resolution image synthesis with latent diffusion models.
\newblock In \emph{CVPR}, 2022.

\bibitem[Ros et~al.(2016)Ros, Sellart, Materzynska, Vazquez, and Lopez]{synthia}
German Ros, Laura Sellart, Joanna Materzynska, David Vazquez, and Antonio~M. Lopez.
\newblock The synthia dataset: A large collection of synthetic images for semantic segmentation of urban scenes.
\newblock In \emph{CVPR}, 2016.

\bibitem[Ruiz et~al.(2023)Ruiz, Li, Jampani, Pritch, Rubinstein, and Aberman]{ruiz2023dreambooth}
Nataniel Ruiz, Yuanzhen Li, Varun Jampani, Yael Pritch, Michael Rubinstein, and Kfir Aberman.
\newblock Dreambooth: Fine tuning text-to-image diffusion models for subject-driven generation.
\newblock In \emph{CVPR}, 2023.

\bibitem[Sakaridis et~al.(2021)Sakaridis, Dai, and Van~Gool]{acdc}
Christos Sakaridis, Dengxin Dai, and Luc Van~Gool.
\newblock Acdc: The adverse conditions dataset with correspondences for semantic driving scene understanding.
\newblock In \emph{ICCV}, 2021.

\bibitem[Sandler et~al.(2018)Sandler, Howard, Zhu, Zhmoginov, and Chen]{sandler2018mobilenetv2}
Mark Sandler, Andrew Howard, Menglong Zhu, Andrey Zhmoginov, and Liang-Chieh Chen.
\newblock Mobilenetv2: Inverted residuals and linear bottlenecks.
\newblock In \emph{CVPR}, 2018.

\bibitem[Shen et~al.(2023)Shen, Gurram, Liu, Wang, and Knoll]{shen2023diga}
Fengyi Shen, Akhil Gurram, Ziyuan Liu, He Wang, and Alois Knoll.
\newblock Diga: Distil to generalize and then adapt for domain adaptive semantic segmentation.
\newblock In \emph{CVPR}, 2023.

\bibitem[Tang et~al.(2021)Tang, Gao, Zhu, Zhang, Li, and Metaxas]{tang2021crossnorm}
Zhiqiang Tang, Yunhe Gao, Yi Zhu, Zhi Zhang, Mu Li, and Dimitris~N Metaxas.
\newblock Crossnorm and selfnorm for generalization under distribution shifts.
\newblock In \emph{ICCV}, 2021.

\bibitem[Udupa et~al.(2024)Udupa, Gurunath, Sikdar, and Sundaram]{udupa2024mrfp}
Sumanth Udupa, Prajwal Gurunath, Aniruddh Sikdar, and Suresh Sundaram.
\newblock Mrfp: Learning generalizable semantic segmentation from sim-2-real with multi-resolution feature perturbation.
\newblock In \emph{CVPR}, 2024.

\bibitem[Vaswani et~al.(2017)Vaswani, Shazeer, Parmar, Uszkoreit, Jones, Gomez, Kaiser, and Polosukhin]{transformer}
Ashish Vaswani, Noam Shazeer, Niki Parmar, Jakob Uszkoreit, Llion Jones, Aidan~N Gomez, \L~ukasz Kaiser, and Illia Polosukhin.
\newblock Attention is all you need.
\newblock In \emph{NeurIPS}, 2017.

\bibitem[Vidit et~al.(2023)Vidit, Engilberge, and Salzmann]{vidit2023clipthegap}
Vidit Vidit, Martin Engilberge, and Mathieu Salzmann.
\newblock Clip the gap: A single domain generalization approach for object detection.
\newblock In \emph{CVPR}, 2023.

\bibitem[von Platen et~al.(2022)von Platen, Patil, Lozhkov, Cuenca, Lambert, Rasul, Davaadorj, Nair, Paul, Berman, Xu, Liu, and Wolf]{von-platen-etal-2022-diffusers}
Patrick von Platen, Suraj Patil, Anton Lozhkov, Pedro Cuenca, Nathan Lambert, Kashif Rasul, Mishig Davaadorj, Dhruv Nair, Sayak Paul, William Berman, Yiyi Xu, Steven Liu, and Thomas Wolf.
\newblock Diffusers: State-of-the-art diffusion models.
\newblock \url{https://github.com/huggingface/diffusers}, 2022.

\bibitem[Wei et~al.(2023)Wei, Chen, Tu, Ling, Chen, and Jin]{dtp}
Zhixiang Wei, Lin Chen, Tao Tu, Pengyang Ling, Huaian Chen, and Yi Jin.
\newblock Disentangle then parse: Night-time semantic segmentation with illumination disentanglement.
\newblock In \emph{ICCV}, 2023.

\bibitem[Wei et~al.(2024)Wei, Chen, Jin, Ma, Liu, Ling, Wang, Chen, and Zheng]{rein}
Zhixiang Wei, Lin Chen, Yi Jin, Xiaoxiao Ma, Tianle Liu, Pengyang Ling, Ben Wang, Huaian Chen, and Jinjin Zheng.
\newblock Stronger fewer \& superior: Harnessing vision foundation models for domain generalized semantic segmentation.
\newblock In \emph{CVPR}, 2024.

\bibitem[Woo et~al.(2023)Woo, Debnath, Hu, Chen, Liu, Kweon, and Xie]{convnextv2}
Sanghyun Woo, Shoubhik Debnath, Ronghang Hu, Xinlei Chen, Zhuang Liu, In~So Kweon, and Saining Xie.
\newblock Convnext v2: Co-designing and scaling convnets with masked autoencoders.
\newblock In \emph{CVPR}, 2023.

\bibitem[Wu and Deng(2022)]{wu2022single-dgod}
Aming Wu and Cheng Deng.
\newblock Single-domain generalized object detection in urban scene via cyclic-disentangled self-distillation.
\newblock In \emph{CVPR}, 2022.

\bibitem[Wu et~al.(2022)Wu, Wu, Zhang, Ju, and Wang]{wu2022siamdoge}
Zhenyao Wu, Xinyi Wu, Xiaoping Zhang, Lili Ju, and Song Wang.
\newblock Siamdoge: Domain generalizable semantic segmentation using siamese network.
\newblock In \emph{ECCV}, 2022.

\bibitem[Xu et~al.(2022)Xu, Yao, Jiang, Jiang, Chu, Han, Zhang, Wang, and Tai]{xu2022dirl}
Qi Xu, Liang Yao, Zhengkai Jiang, Guannan Jiang, Wenqing Chu, Wenhui Han, Wei Zhang, Chengjie Wang, and Ying Tai.
\newblock Dirl: Domain-invariant representation learning for generalizable semantic segmentation.
\newblock In \emph{AAAI}, 2022.

\bibitem[Yang et~al.(2023)Yang, Gu, and Sun]{dpcl}
Liwei Yang, Xiang Gu, and Jian Sun.
\newblock Generalized semantic segmentation by self-supervised source domain projection and multi-level contrastive learning.
\newblock In \emph{AAAI}, 2023.

\bibitem[Yi et~al.(2024)Yi, Bi, Zheng, Zhan, Ji, Huang, Li, and Zheng]{set}
Jingjun Yi, Qi Bi, Hao Zheng, Haolan Zhan, Wei Ji, Yawen Huang, Yuexiang Li, and Yefeng Zheng.
\newblock Learning spectral-decomposited tokens for domain generalized semantic segmentation.
\newblock In \emph{ACMMM}, 2024.

\bibitem[Yu et~al.(2020)Yu, Chen, Wang, Xian, Chen, Liu, Madhavan, and Darrell]{bdd}
Fisher Yu, Haofeng Chen, Xin Wang, Wenqi Xian, Yingying Chen, Fangchen Liu, Vashisht Madhavan, and Trevor Darrell.
\newblock Bdd100k: A diverse driving dataset for heterogeneous multitask learning.
\newblock In \emph{CVPR}, 2020.

\bibitem[Yu et~al.(2024)Yu, Shin, Back, Ko, Noh, and Lee]{dplot}
Yeonguk Yu, Sungho Shin, Seunghyeok Back, Mihwan Ko, Sangjun Noh, and Kyoobin Lee.
\newblock Domain-specific block selection and paired-view pseudo-labeling for online test-time adaptation.
\newblock In \emph{CVPR}, 2024.

\bibitem[Yue et~al.(2019)Yue, Zhang, Zhao, Sangiovanni-Vincentelli, Keutzer, and Gong]{randomization}
Xiangyu Yue, Yang Zhang, Sicheng Zhao, Alberto Sangiovanni-Vincentelli, Kurt Keutzer, and Boqing Gong.
\newblock Domain randomization and pyramid consistency: Simulation-to-real generalization without accessing target domain data.
\newblock In \emph{ICCV}, 2019.

\bibitem[Zhao et~al.(2022)Zhao, Zhong, Zhao, Sebe, and Lee]{shade}
Yuyang Zhao, Zhun Zhong, Na Zhao, Nicu Sebe, and Gim~Hee Lee.
\newblock Style-hallucinated dual consistency learning for domain generalized semantic segmentation.
\newblock In \emph{ECCV}, 2022.

\bibitem[Zhong et~al.(2022)Zhong, Zhao, Lee, and Sebe]{advstyle}
Zhun Zhong, Yuyang Zhao, Gim~Hee Lee, and Nicu Sebe.
\newblock Adversarial style augmentation for domain generalized urban-scene segmentation.
\newblock In \emph{NeurIPS}, 2022.

\bibitem[Zhou et~al.(2021)Zhou, Wei, Wang, Shen, Xie, Yuille, and Kong]{zhou2021ibot}
Jinghao Zhou, Chen Wei, Huiyu Wang, Wei Shen, Cihang Xie, Alan Yuille, and Tao Kong.
\newblock ibot: Image bert pre-training with online tokenizer.
\newblock \emph{arXiv:2111.07832}, 2021.

\bibitem[Zong et~al.(2023)Zong, Song, and Liu]{codetr}
Zhuofan Zong, Guanglu Song, and Yu Liu.
\newblock Detrs with collaborative hybrid assignments training.
\newblock In \emph{ICCV}, 2023.

\end{thebibliography}
}

\end{document}